\pdfoutput=1
\documentclass[11pt]{article}

\PassOptionsToPackage{hyperfootnotes=false}{hyperref}
\usepackage[final]{styles/acl}
\usepackage{times}
\usepackage{latexsym}
\usepackage[T1]{fontenc}
\usepackage[utf8]{inputenc}
\usepackage{microtype}
\usepackage{inconsolata}
\usepackage{graphicx}
\usepackage{amsmath}
\usepackage{booktabs}
\usepackage{multirow}
\usepackage{tabularx}
\usepackage{xspace}
\usepackage{xcolor}
\usepackage{algorithm}
\usepackage{algpseudocode}
\usepackage{placeins}
\usepackage{float}
\usepackage{pgfplots}
\pgfplotsset{compat=1.18}
\usepackage{changepage}
\usepackage{wrapfig}
\usepackage[font=footnotesize,labelfont=bf,skip=1pt]{caption}
\usepackage{array}
\usepackage[most]{tcolorbox}
\newtcblisting{promptbox}{
  colback=gray!5,
  colframe=gray!40,
  boxrule=0.4pt,
  arc=1pt,
  left=4pt, right=4pt, top=3pt, bottom=3pt,
  listing only,
  listing options={
    basicstyle=\scriptsize\ttfamily,
    breaklines=true,
    breakautoindent=false,
    columns=fullflexible,
    aboveskip=0pt,
    belowskip=0pt,
  },
  breakable,
}


\newcommand{\method}{TAG-Nav\xspace}

\newcommand{\benchmark}{DestBench\xspace}
\newcommand{\appicon}[1]{\includegraphics[height=1.2em,keepaspectratio]{#1}}

\title{Not All Graphs Are Maps: A Task-Agnostic Exploration Framework for Mobile GUI Navigation}

\author{\normalfont
\textbf{Adamo Orsini}\textsuperscript{1,2}\thanks{These authors contributed equally.},
\textbf{Seyed Mehdi Ayyoubzadeh}\textsuperscript{3}\footnotemark[1],
\textbf{Sadman Rakib Pinon}\textsuperscript{1,2}\\
\textbf{Hannuo Zhang}\textsuperscript{1,2},
\textbf{Koosha Gholipour Baradari}\textsuperscript{1},
\textbf{Yuanhao Yu}\textsuperscript{3}\\
\textbf{Xinxin Zuo}\textsuperscript{1}, \textbf{Yang Wang}\textsuperscript{1,2}\\
{\normalsize\textsuperscript{1}Concordia University \quad \textsuperscript{2}Mila -- Quebec AI Institute \quad \textsuperscript{3}McMaster University}\\
{\footnotesize\texttt{\{adamo.orsini,sadmanrakib.pinon,hannuo.zhang,koosha.gholipourbaradari\}@mail.concordia.ca}}\\
{\footnotesize\texttt{s.mehdi.ayyoubzadeh@gmail.com}\quad\texttt{dr.yhyu@outlook.com}}\\
{\footnotesize\texttt{\{xinxin.zuo,yang.wang\}@concordia.ca}}
}

\begin{document}
\raggedbottom
\maketitle

\begin{abstract}
Modern mobile GUI agents can perceive visible screens and execute local actions, but they still lack persistent app-specific navigation knowledge: where functions live, which pages contain them, and which screens are equivalent across repeated visits.
We introduce \method, a Task-Agnostic Graph Navigation framework that wraps existing GUI-agent backbones with reusable destination memory.
Offline exploration is distilled into compact records of app functions, target pages, and waypoint routes before any evaluated instruction is seen.
When a future task's destination is covered, \method retrieves the destination and provides route guidance without task-specific demonstrations.
This separates mobile GUI \emph{destination navigation} from full task completion: many tasks share the prefix of reaching the right page, while final actions may enter content, send data, make purchases, delete information, or change sensitive settings.
To evaluate route finding without downstream execution confounds, we introduce \benchmark, a navigation-only protocol that asks whether an agent reaches the correct pre-action destination screen.
\benchmark enables a controlled graph-memory study comparing no-memory agents with random, automated BFS, and clean graph memory.
Graph-derived memory improves strict navigation success rate of UI-TARS-1.5-7B from 52.4\% without graph memory to 58.3\% with automated BFS memory, and improves Qwen-VL-Max success rate from 59.5\% to 71.4\% on the same conditions. 
Together, \method and \benchmark\footnote{\url{https://tag-nav-authors.github.io/tag_nav/}} show that destination coverage and route cleanliness strongly shape graph-memory value, making graph quality a measured variable rather than an assumed benefit of exploration.
\end{abstract}

\section{Introduction}
\label{sec:introduction}

Modern multimodal GUI agents perceive screens and execute visible actions well, but they lack a persistent map of the apps they operate \citep{zhang2023appagent,wen2024autodroid,qin2025uitars}.
When asked to reach a deep setting, account-management page, privacy control, or app-specific tool, the agent must re-infer where the function lives and how to navigate there from the current screen on every task.
Many user requests share the prefix of arriving at a destination before any task-specific (and often sensitive or irreversible) final action: reaching an account-privacy screen is reusable navigation, while toggling the control itself is task-specific execution.

We study mobile GUI \emph{destination navigation} as a problem distinct from full task completion, focused on whether the agent reaches the correct pre-action destination screen.

The standard response is to explore the app offline and retrieve from the resulting graph \citep{wen2024autodroid,guan2025kgrag,xie2025guiexplorer,chen2025pgagent}, but exploration does not automatically yield a useful map: a large random graph can have low target coverage and noisy routes, while a small clean graph can be highly navigable.
Two graph-construction properties decide whether explored memory actually helps: \emph{target coverage} of the destinations users request, and \emph{route cleanliness} from the app root to those destinations.
We argue that graphs satisfying both, built by task-agnostic breadth-oriented exploration and post-processed into destination-anchored deduplicated routes, are the design that turns offline exploration into navigation memory that wraps any visual GUI actor.

We instantiate this design in \method, a \textbf{T}ask-\textbf{A}gnostic \textbf{G}raph \textbf{Nav}igation framework that wraps existing visual GUI agents with destination memory without retraining.
\method explores apps with task-agnostic local breadth-first search (Local-BFS) and post-processes the result into compact destination records (target page summary, intents, waypoint route, transition hints); at inference time the matching record is injected as plain text into an unmodified backbone such as UI-TARS \citep{qin2025uitars}.
The retrieved unit is a destination plus its route, not a per-task trajectory replay, so \method can be evaluated under different graph-construction conditions while holding the actor and task set fixed.

To evaluate this design on real apps, we introduce \benchmark, a navigation-only protocol on real Android apps that judges whether the agent reaches the correct pre-action destination screen, with graph-aware target coverage and route cleanliness diagnostics.
End-to-end task benchmarks would otherwise conflate memory quality with downstream execution, blurring the properties we want to measure.
On 84 \benchmark tasks spanning six apps, \method achieves higher observed navigation success than a no-memory baseline across two GUI backbones (Table~\ref{tab:cross-backbone}). Controlled comparisons against random-exploration and clean manual-graph conditions examine the roles of target coverage and route cleanliness (Table~\ref{tab:graph-quality}).

Our contributions are:
\begin{itemize}
    \setlength{\itemsep}{0.1em}
    \setlength{\topsep}{0.2em}
    \item \textbf{\method}, a task-agnostic destination-memory module that wraps an unmodified visual GUI actor: offline exploration becomes compact destination records (target page, waypoint route, transition hints) injected as plain text at inference.
    \item \textbf{\benchmark}, a navigation-only page-reaching protocol on real Android apps with graph-aware diagnostics for target coverage and route cleanliness.
\end{itemize}

\section{Related Work}
\label{sec:related}

\paragraph{Mobile GUI Agents and Visual Grounding.} Mobile GUI agents translate natural language instructions and visual observations into interface actions.
Smartphone agents and Android automation systems showed that language models can operate mobile apps through screenshots, UI trees, and action histories \citep{zhang2023appagent,wen2024autodroid,wang2024mobileagentv2,wang2025mobileagente}.
More recent general-purpose GUI backbones such as CogAgent and UI-TARS improve perception, grounding, and multi-step action generation across GUI tasks \citep{hong2024cogagent,qin2025uitars,qin2025uitars2}.
Parallel work on visual grounding and screen parsing, including SeeClick, UGround, and OmniParser, shows that locating the correct on-screen element is itself a central bottleneck for visual GUI agents \citep{cheng2024seeclick,gou2025uground,lu2024omniparser}.
\method is complementary to these efforts: it does not replace the actor or the grounding model, but supplies app-specific destination memory that tells an existing actor where to go.
Table~\ref{tab:related} summarizes how \method differs from prior graph- and retrieval-based systems on two axes that anchor the rest of this section: \emph{what is retrieved online} and \emph{whether the memory wraps an unmodified actor}; a per-system breakdown (stored memory, retrieved unit, coupling) is given in Appendix~\ref{app:related-positioning}.

\begin{table}[t]
\centering
\footnotesize
\setlength{\tabcolsep}{2pt}
\renewcommand{\arraystretch}{1.2}
\begin{tabularx}{\columnwidth}{@{}>{\raggedright\arraybackslash}p{0.30\columnwidth}>{\raggedright\arraybackslash}X>{\centering\arraybackslash}p{0.10\columnwidth}@{}}
\toprule
\textbf{Systems} & \textbf{What is retrieved online} & \textbf{Actor} \\
\midrule
AutoDroid, AppAgent \citep{wen2024autodroid,zhang2023appagent} & Per-element hints or\newline app docs & No \\
KG-RAG, GUI-explorer \citep{guan2025kgrag,xie2025guiexplorer} & Trajectory pairs or current-screen cues & No \\
PG-Agent, GraphPilot \citep{chen2025pgagent,yu2026graphpilot} & Page-graph guidelines or one-shot plans & No \\
\midrule
\textbf{\method} & \textbf{Destination page + waypoint route} & \textbf{Yes} \\
\bottomrule
\end{tabularx}
\caption{\textbf{Positioning of \method against representative graph- and memory-augmented GUI agents.} Two axes capture the main difference: the \emph{unit of memory} retrieved online, and whether that memory wraps an \emph{unmodified} stepwise visual actor. Detailed per-system positioning (exploration mode, graph structure, actor pipeline) is deferred to Appendix~\ref{app:related-positioning}.}
\label{tab:related}
\end{table}

\begin{figure*}[!t]
  \centering
  \includegraphics[
    width=\linewidth,
    trim=0 0 0 0,
    clip
  ]{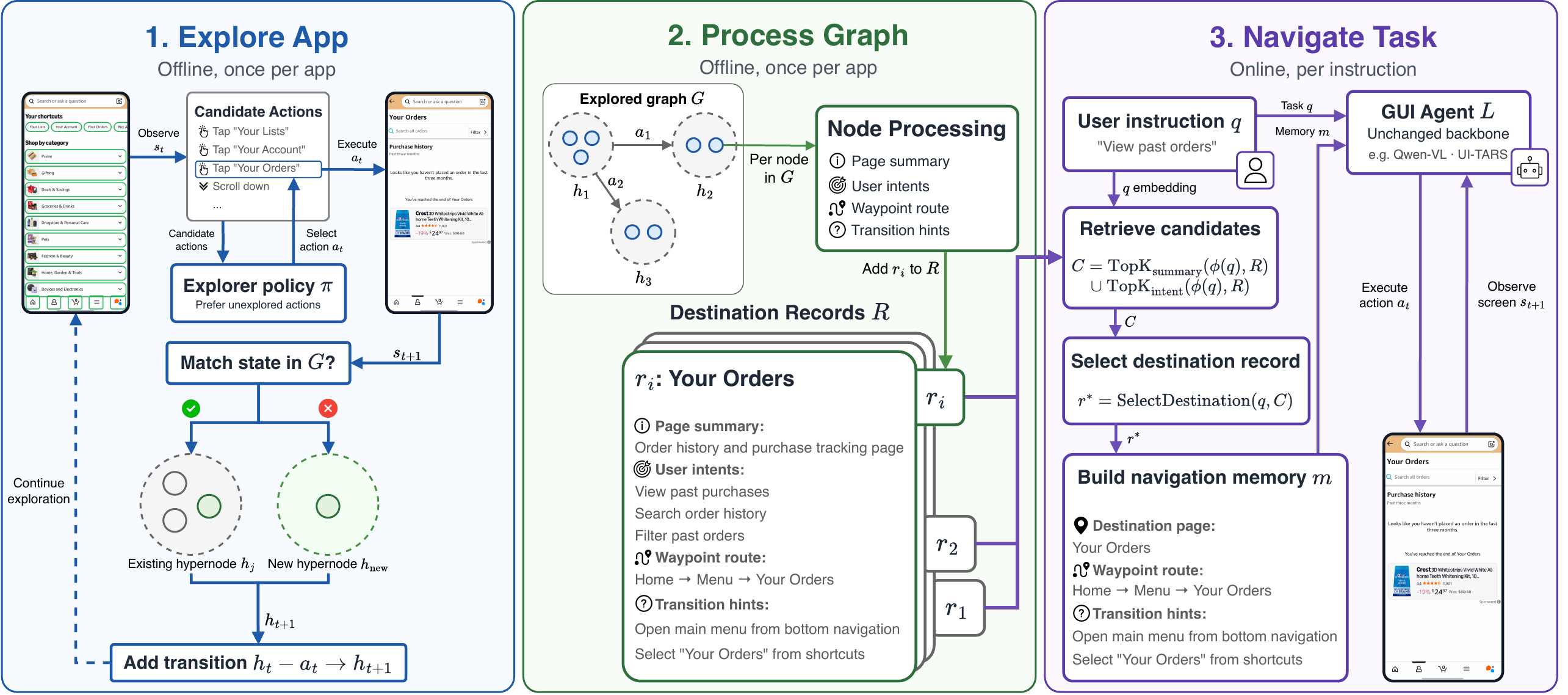}
  \caption{\textbf{Overview of \method on an Amazon instruction}. Offline exploration builds a hypernode app graph, post-processing distills indexed states into destination records, and online navigation retrieves candidate records, selects $r^\ast$, and formats the selected record as destination memory for an unchanged GUI agent. The figure writes $r_i$ for the formal record $r_{h,n}$ with $i=(h,n)$ and uses $s_t$ only as schematic exploration-state notation; the formal online/evaluation trajectory uses screenshots $x_t$.}
  \label{fig:pipeline}
\end{figure*}

\paragraph{Mobile-Agent Benchmarks.} Mobile-agent benchmarks have expanded from UI-control datasets to interactive task-completion environments.
Android in the Wild and AndroidControl study large-scale supervised UI-control data with high- and low-level instructions \citep{rawles2023aitw,li2024androidcontrol}, while AndroidLab provides a reproducible Android training and evaluation framework \citep{xu2024androidlab}.
AndroidWorld, SPA-Bench, Mobile-Bench, and MobileWorld evaluate increasingly realistic mobile-agent task completion, including dynamic environments, multi-step planning, API/tool use, and user interaction \citep{rawles2024androidworld,chen2024spabench,deng2024mobilebench,kong2025mobileworld}.
Other recent benchmarks stress cross-app navigation, multiple valid paths, noisy apps, and online procedural judging \citep{lu2025guiodyssey,xu2025mobilebenchv2,im2025mobibench,chai2025a3}.
These benchmarks are valuable, but their success criteria usually mix navigation with downstream execution: typing user-specific content, changing app state, sending information, selecting items, or satisfying procedural completion checks.
Because they sample task completion across many apps, they also provide limited visibility into how well app-specific memory covers the functional regions of a single target app.
\benchmark instead isolates page-reaching navigation by asking whether the agent reaches the pre-action destination screen for a given task.
This makes it possible to evaluate graph-derived navigation memory through target coverage, target retrieval, path existence, route cleanliness, and strict page matching rather than only final task success.

\paragraph{App Exploration and App-Specific Memory.} Several systems build app-specific knowledge before task execution.
Earlier UI-guided testing tools such as DroidBot built app state-transition models through device interaction, providing an important predecessor to app exploration for automation \citep{li2017droidbot}.
AutoDroid explores Android apps, builds a UI transition graph, and turns view-hierarchy/accessibility states into simplified HTML prompts and simulated-task app memory \citep{wen2024autodroid}.
Its memory is structured around explored UI trees, elements, and actions rather than visual screen perception, making it a strong early example of exploration-based app memory.
AppAgent and AppAgent-v2 use autonomous or manual exploration to build app documents for later task execution \citep{zhang2023appagent,zhang2024appagentv2}.
Their memory is element- and document-centered: per-element notes consulted on the current screen, not destination memory keyed to a target page.
GUI-Xplore uses pre-recorded exploration videos and converts them into GUI transition graphs \citep{sun2025guixplore}.
GUI-Xplore is primarily evaluated as an offline app-understanding setup over those videos, whereas \method studies graph memory under live device interaction with a stepwise actor.
GUI-explorer autonomously explores apps and mines transition-aware knowledge for dynamic guidance \citep{xie2025guiexplorer}.
Across this family, exploration-derived memory is largely treated as an assumed benefit; what varies between systems is the exploration policy and the memory format, not a controlled comparison of when the explored graph yields useful navigation memory.
Our narrower question is when exploration becomes reusable navigation memory: whether the graph contains destination pages, whether retrieval selects the right target, and whether the recovered route is clean enough to guide a future page-reaching task.

\begin{figure*}[!t]
\centering
\includegraphics[width=0.95\linewidth]{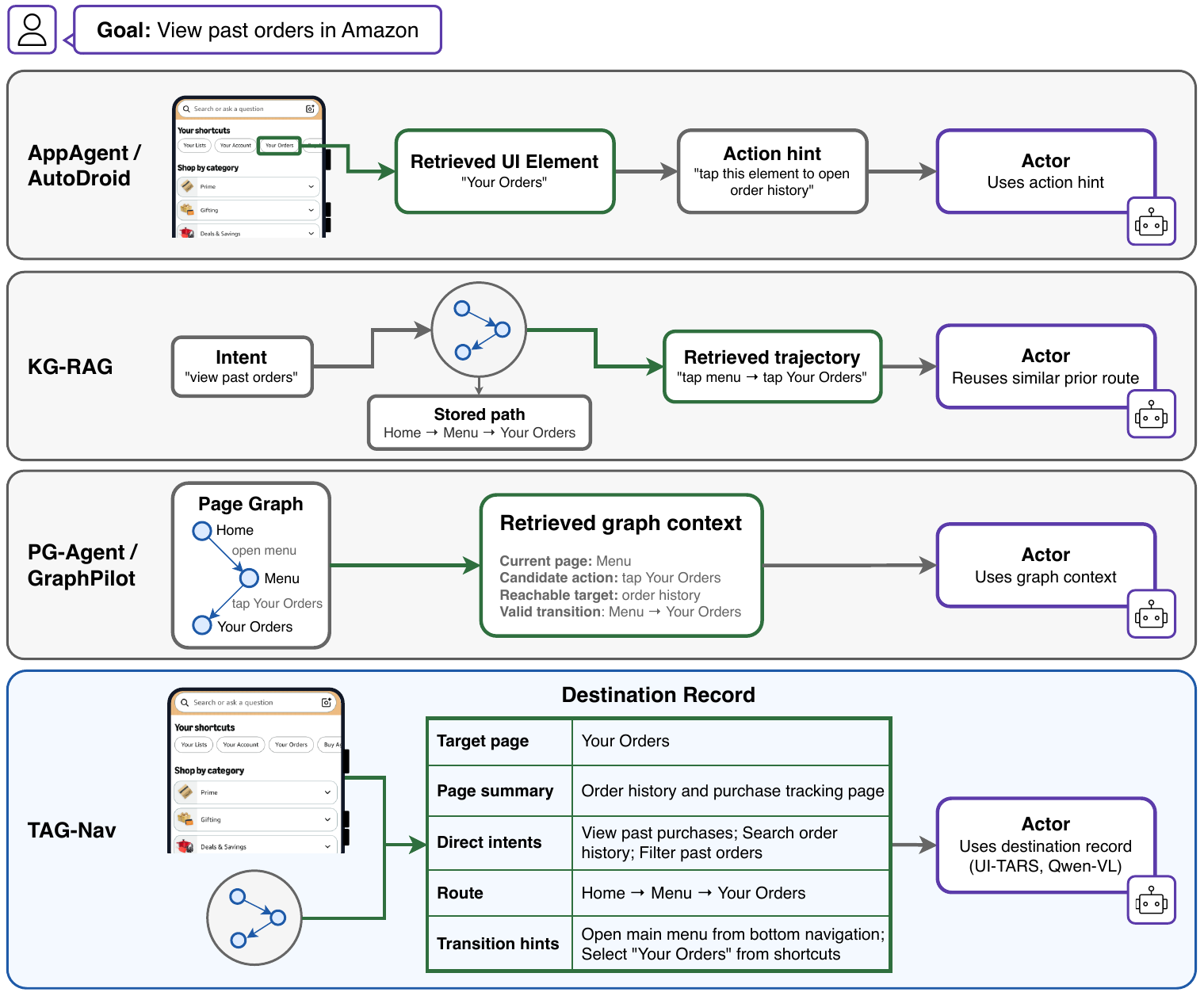}
\caption{\textbf{Memory units retrieved at inference by representative graph- and memory-augmented GUI agents.} AutoDroid and AppAgent retrieve a per-element hint anchored to a specific UI control; KG-RAG retrieves an intent-trajectory pair (a flat replay of how a similar past task was completed); PG-Agent and GraphPilot retrieve page-level guideline text. \method\ instead retrieves a structured \emph{destination record} with a named target, page summary, route, and transition hints, and injects it into an unmodified visual actor (UI-TARS, Qwen-VL). The categorical difference between \method\ and prior work is the \emph{shape} of the retrieved memory unit, not the presence of memory.}
\label{fig:memory-unit-comparison}
\end{figure*}

\paragraph{Graph and Retrieval-Augmented GUI Agents.} Graph and retrieval memories are increasingly common in GUI-agent systems.
KG-RAG transforms UI transition graphs into a vector database of intent-trajectory pairs and retrieves relevant trajectories at inference time \citep{guan2025kgrag}.
PG-Agent reconstructs page graphs from episodes and retrieves page-graph guidelines for multi-agent planning \citep{chen2025pgagent}.
GraphPilot constructs app-specific knowledge graphs and uses them for low-latency action-sequence generation \citep{yu2026graphpilot}.
HyMEM studies graph-structured self-evolving memory for GUI agents \citep{zhu2026hymem}, while AndroTMem focuses on anchored within-task memory in long-horizon Android tasks \citep{shi2026androtmem}.
The closest systems differ in what memory they retrieve and how it is consumed: KG-RAG retrieves intent-trajectory pairs for action decisions, GUI-explorer retrieves transition-aware current-screen knowledge for dynamic guidance, PG-Agent retrieves screen-conditioned page-graph guidelines for multi-agent planning, and GraphPilot uses a knowledge graph to plan near one-shot action sequences with a validator.
\method differs along two axes that prior work does not exercise together (see Fig.~\ref{fig:memory-unit-comparison}): the online unit is a destination page with a compact waypoint route consumed by an unmodified stepwise visual actor, and graph quality is measured by comparing no-memory, random, automated-BFS, and clean-graph conditions under the same actor and tasks.

\section{\method}
\label{sec:method}

\subsection{Overview}

\method is a navigation-memory module for mobile GUI agents that separates app-specific discovery from online task execution.
Offline exploration builds an app graph over hypernode-clustered observations, then post-processing turns each indexed graph state into a compact destination record describing what the page is for and how it can be reached.
At inference time, \method retrieves the destination record for the user instruction, formats it as a plain-text destination-memory block, and injects it into an existing GUI-agent backbone without training a new actor or replacing the actor's visual grounding module.
Figure~\ref{fig:pipeline} summarizes the full pipeline: the offline input is an app/device configuration and the offline output is an app graph plus target-indexed records; the online input is a natural-language instruction and the online output is the executed trajectory together with the evaluator's terminal status.

\subsection{Offline App Exploration}
\label{sec:offline-exploration}

\method represents an app as a directed graph $G=(\mathcal{H},E)$.
A hypernode $h \in \mathcal{H}$ groups observed UI states that the matcher judges to be the same screen or screen family using visual or layout evidence.
Each member observation stores its screenshot, parsed UI or layout evidence when available, and metadata collected during exploration.
An edge $e=(h_i,a,h_j)$ records an executable transition from source hypernode $h_i$ to destination hypernode $h_j$, where $a$ includes the action type, an action description, and available element evidence.
The graph is therefore a state abstraction over repeated screenshots rather than a literal log of every frame.

Auto-BFS is our automated graph-construction condition, which builds the app graph through repeated walks from the app's initial screen. It combines walks using the local breadth-first search exploration policy (Local-BFS) with random-action walks. Local-BFS walks probe unexplored actions and verify return to the current screen before probing another action, skipping actions already explored on matched sibling nodes within the same hypernode. Observed states are matched to existing hypernodes or assigned to new ones, and transitions are recorded as edges. We treat Auto-BFS as an autonomous graph-construction condition rather than a novel graph-search algorithm. Appendix~\ref{app:local-bfs} gives the walk schedule and measured exploration counts, and Appendix~\ref{app:hypernode-match} describes how observed states are matched to hypernodes.

\begin{algorithm}[t]
\caption{Offline Graph Memory Construction}
\label{alg:offline}
\footnotesize
\begin{algorithmic}[1]
\Require app $A$, device $D$, exploration configuration $\pi$, maximum walks $N$, VLM $M_{\mathrm{mem}}$, embedder $\phi$
\Ensure graph $G=(\mathcal{H},E)$, memory index $R$
\State $s_0 \gets \mathsf{Observe}(\mathsf{Reset}(A,D))$
\State $h_0 \gets \mathsf{Hypernode}(s_0)$
\State $\mathcal{H} \gets \{h_0\}$; $E \gets \emptyset$; $R \gets \emptyset$
\State $G \gets (\mathcal{H},E)$
\For{$w=1,\ldots,N$}
    \State $s \gets \mathsf{Observe}(\mathsf{Reset}(A,D))$
    \State $h \gets \mathsf{MatchOrAdd}(s,G)$
    \State $m \gets \mathsf{SampleWalkMode}(\pi)$
    \State $D_w \gets \mathsf{DepthLimit}(\pi,w)$; $d \gets 0$; $k \gets 0$
    \While{$d < D_w$}
        \State $(a,b) \gets \mathsf{SelectAction}(s,h,G,m,d,k,\pi)$
        \State $(s',\ell') \gets \mathsf{ExecuteObserve}(D,a)$
        \State $h' \gets \mathsf{MatchHypernode}(s',\ell',G)$
        \If{$h'=\emptyset$}
            \State $h' \gets \mathsf{Hypernode}(s',\ell')$
            \State $\mathcal{H} \gets \mathcal{H} \cup \{h'\}$
        \Else
            \State $\mathsf{AddObservation}(h',s',\ell')$
        \EndIf
        \State $E \gets E \cup \{(h,a,h')\}$
        \State $\mathsf{MarkExplored}(s,h,a,G)$
        \State $\mathsf{Checkpoint}(G,s',a)$
        \If{$b$}
            \State $(s_{\mathrm{back}},v) \gets \mathsf{UndoAndCheck}(D,a,s,h,\pi)$
            \If{$\neg v$}
                \State \textbf{break}
            \EndIf
            \State $s \gets s_{\mathrm{back}}$
        \Else
            \State $(s,h) \gets (s',h')$
        \EndIf
        \State Update walk counters $(d,k)$ according to $\pi$
    \EndWhile
\EndFor
\ForAll{$h \in \mathcal{H}$, $n \in \mathsf{Members}(h)$}
    \State $P_{h,n} \gets \mathsf{RootPaths}(G,h,n)$
    \State $(d_{h,n},I_{h,n},W_{h,n},H_{h,n}) \gets M_{\mathrm{mem}}(h,n,P_{h,n})$
    \State $z^d_{h,n} \gets \phi(d_{h,n})$; $z^I_{h,n} \gets \phi(I_{h,n})$
    \State $R[(h,n)] \gets$
    \Statex \hspace{\algorithmicindent}\hspace{\algorithmicindent}$(h,n,d_{h,n},I_{h,n},W_{h,n},H_{h,n},z^d_{h,n},z^I_{h,n})$
\EndFor
\State \Return $G, R$
\end{algorithmic}
\end{algorithm}

In Algorithm~\ref{alg:offline}, $\pi$ specifies the exploration configuration described in Appendix~\ref{app:local-bfs}, including the probability of local expansion, depth schedule, maximum number of successful undo-and-check cycles, and rules for selecting actions. The counters $d$ and $k$ track walk depth and successful undo-and-check cycles, respectively. $\mathsf{SelectAction}$ returns an action and a flag $b$ indicating whether it is a local probe. Local-BFS skips actions already explored on the current screen or matching actions explored on sibling screens within the same hypernode. $\mathsf{UndoAndCheck}$ returns the observed landing state and a flag $v$ indicating whether return to the source screen was verified after any configured recovery attempts. $\mathsf{MatchOrAdd}$ matches the reset observation or adds it to $G$; graph updates modify $\mathcal{H}$ and $E$ in place.

\subsection{Graph Construction and Post-Processing}

Raw exploration graphs are not used directly as prompts: mobile apps contain dynamic content, ads, pop-ups, scroll variants, and alternate paths that would produce noisy targets and brittle action scripts if copied verbatim.
\method therefore post-processes the graph into compact retrieval records that describe what each destination screen is for and how it can be reached.

For each indexed graph state, a vision-language model generates a concise visual page summary and a set of user intents directly satisfied by the current screen.
The prompt emphasizes the active interaction layer of the final screenshot, so if a modal, drawer, menu, or permission dialog is in the foreground, the metadata describes that foreground layer rather than the dimmed background.

\method also constructs the route field of each destination-memory record by extracting root-to-state paths from the graph, annotating the screenshots along each path with action evidence, and asking a vision-language model to distill the route into semantic waypoints and transition hints.
The post-processing prompt removes loops, retries, oscillations, and irrelevant detours, so the resulting memory is expressed as page or state labels and high-level transitions, not fixed coordinates or event-log replay.

Each processed graph state $(h,n)$ becomes a memory record $r_{h,n}=(h,n,d_{h,n},I_{h,n},W_{h,n},H_{h,n},z^d_{h,n},z^I_{h,n})$ with the summary, intent, waypoint, hint, and embedding fields defined in Algorithm~\ref{alg:offline}; we write $i=(h,n)$ and $r_i=r_{h,n}$ for compact figure notation.
Figure~\ref{fig:memory-unit-comparison} contrasts the destination-record unit with the per-element hints, intent-trajectory pairs, and page guidelines used by prior graph-RAG GUI systems \citep{wen2024autodroid,guan2025kgrag,xie2025guiexplorer,chen2025pgagent}.

\subsection{Target Retrieval and Memory Formatting}

Given a user instruction $q$, \method first retrieves candidate destination records.
The instruction is embedded and compared against both the visual-summary embeddings and direct-intent embeddings for all records.
Top matches from the two fields are combined so that a candidate may be retrieved because it looks like the requested page, because it directly supports the requested intent, or both.

The candidate set is then passed to a compact target-selection prompt.
The selector receives the user instruction, candidate screen summaries, and direct-intent text, then chooses the graph state whose current screen best satisfies the instruction as the final destination.
The selector prefers loaded destinations over loading or transition screens, the destination page over foreground modals unless requested, routes page-style queries to the visual summary and action-style queries to the intents, rejects link or menu entries that merely point at the target, and prefers specific destinations for specific queries and general help pages for generic queries (full prompts in Appendix~\ref{app:prompts}).
This target-first retrieval step avoids a common graph-memory failure mode in which a retrieved state is semantically close to the task but is not itself the destination page; unlike a direct retrieval of a past task trace, the destination is selected first and its memory is only then formatted for injection.

\method then formats a destination-memory block containing the original task, a final-goal description, the relevant waypoint sequence, and transition hints.
The block is intentionally semantic and route-oriented: it orients the backbone toward the correct app region while leaving grounding, action selection, and local recovery to the backbone, and it never prescribes fixed coordinates, replays exploration verbatim, or asks the actor to perform the irreversible final task action.

\subsection{Agent-Agnostic Online Guidance}

Algorithm~\ref{alg:online} formalizes the online actor loop: \method launches the app, captures the current screenshot, calls the GUI backbone with the user instruction plus the retrieved destination-memory block, and executes the predicted action until the backbone emits a terminal \texttt{finished} action or the step budget is reached.

The memory interface itself is backbone-independent: the destination-memory block is plain text, and any GUI agent that consumes a task prompt and screen observation and emits mobile actions can be wrapped with the same retrieved memory without retraining. The canonical backbone in our experiments is UI-TARS-1.5-7B; Sec.~\ref{sec:experiments} pairs \method memory with different GUI actors to test this agent-agnostic property empirically. Dynamic variants that refresh or edit memory during execution are treated as controlled ablations in Appendix~\ref{app:dynamic-variants}.

\begin{algorithm}[t]
\caption{Online Graph-Memory Navigation}
\label{alg:online}
\footnotesize
\begin{algorithmic}[1]
\Require instruction $q$, memory index $R$, embedder $\phi$, backbone $L$, device $D$, budget $T$
\Ensure executed trajectory $\tau$, terminal status $y$
\State $C_{\mathrm{sum}} \gets \mathsf{TopK}_{\mathrm{summary}}(\phi(q),R)$
\State $C_{\mathrm{int}} \gets \mathsf{TopK}_{\mathrm{intent}}(\phi(q),R)$
\State $C \gets C_{\mathrm{sum}} \cup C_{\mathrm{int}}$
\State $r^\ast \gets \mathsf{SelectDestination}(q,C)$
\State $m \gets \mathsf{FormatMemory}(q,r^\ast)$
\State $\tau \gets \emptyset$
\For{$t=1,\ldots,T$}
    \State $x_t \gets \mathsf{Screenshot}(D)$
    \State $a_t \gets L(q,x_t,m)$
    \State $\tau \gets \mathsf{Append}(\tau,(x_t,a_t))$
    \If{$a_t=\textsc{Finished}$}
        \State $y \gets \textsc{Finished}$
        \State \Return $\tau, y$
    \EndIf
    \State $\mathsf{Execute}(D,a_t)$
\EndFor
\State $y \gets \textsc{StepBudgetReached}$
\State \Return $\tau, y$
\end{algorithmic}
\end{algorithm}

\section{\benchmark}
\label{sec:destbench}

End-to-end task benchmarks conflate navigation with grounding, manipulation, and finalization, so an aggregate success rate cannot attribute a failure to memory.

\subsection{Page-Reaching Navigation}

Each \benchmark instance is a tuple $(A, q, s_0, x_{\mathrm{target}})$ where $A$ is the target app and version, $q$ is a natural-language instruction, $s_0$ is a reset starting state, and $x_{\mathrm{target}}$ is a ground-truth screenshot of the pre-action destination screen. The agent emits a trajectory $\tau = (x_0, a_0, x_1, a_1, \ldots, x_T)$ that terminates when it issues a \textsc{Finished} action or hits the step budget. Success is judged by visual equivalence between $x_T$ and $x_{\mathrm{target}}$ rather than by completion of any downstream operation: a task like ``set an alarm for 7am'' is reduced to ``reach the new-alarm creation screen''; the time picker itself is out of scope. This is a deliberate measurement choice rather than a simplification of mobile use: the post-arrival action is where benchmarks introduce confounds (form-filling, sensitive state changes, tool or API use) that have little to do with app-specific navigation. Accordingly, DestBench scores page reaching only; its gains should not be interpreted as gains in end-to-end task completion.

\subsection{Apps, Tasks, and Targets}

\benchmark contains tasks from 20 Android apps spanning Shopping, Productivity, Utility, Travel, Entertainment, Social Media, Food \& Drink, Communication, and Local Services categories (full inventory in Appendix~\ref{app:app-selection}).
The current controlled graph-memory experiments use the 6-app core cohort for which target annotations and graph conditions are available.
Tasks are derived from real user trajectories collected in our benchmark substrate, with each task annotated by the pre-action destination screen from its original trajectory (annotation protocol in Appendix~\ref{app:annotation}); Table~\ref{tab:destbench-task-distribution} summarizes per-app task counts.

\begin{table}[H]
\centering
\scriptsize
\setlength{\tabcolsep}{3pt}
\renewcommand{\arraystretch}{1.05}
\begin{tabular}{@{}>{\centering\arraybackslash}m{1.2em}lcrrrr@{}}
\toprule
 & \textbf{App} & \textbf{Tasks} & \textbf{Records} & \textbf{Hypernodes} & \textbf{Edges} & \textbf{Covered} \\
\midrule
\appicon{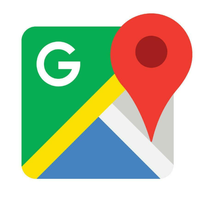} & Google Maps & 24 & 415 & 364 & 1,192 & 20 \\
\appicon{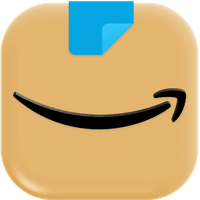}      & Amazon      & 17 & 287 & 257 & 652 & 8 \\
\appicon{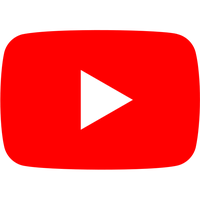}     & YouTube     & 15 & 229 & 196 & 483 & 9 \\
\appicon{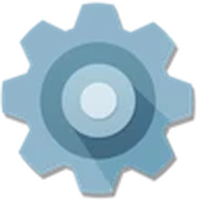}    & Settings    & 9 & 94 & 84 & 128 & 7 \\
\appicon{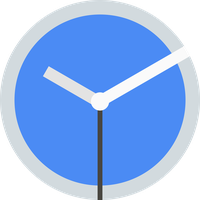}       & Clock       & 9 & 93 & 91 & 322 & 6 \\
\appicon{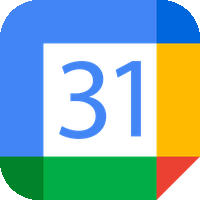}    & Calendar    & 10 & 50 & 40 & 182 & 5 \\
\midrule
 & Total & 84 & 1,168 & 1,032 & 2,959 & 55 \\
\bottomrule
\end{tabular}
\caption{\textbf{Per-app task distribution and core graphs in \benchmark.} The controlled graph-memory experiments use the six-app core cohort shown here. Graph columns describe the Auto-BFS graphs (one record per stored screen state; directed hypernode edges); Covered counts tasks whose strict target is in the graph (human-audited).}
\label{tab:destbench-task-distribution}\label{tab:graph-stats-coverage}
\end{table}

\subsection{Graph-Aware Diagnostics}

\benchmark records per-task structural metadata that attributes failures to specific causes rather than only reporting aggregate success: \emph{target coverage} (whether the graph contains a node visually equivalent to $x_{\mathrm{target}}$), \emph{retrieval correctness} (whether top-1 retrieval picks the correct target), \emph{path existence} from $s_0$ to the retrieved target, \emph{path cleanliness} (route length in waypoints, detours, loops, backtracks, restarts, and transition-hint count), and \emph{judge match score} (raw 1--5 rating). A target is covered only when a record matches the pre-action destination; a parent page or a screen one action too early or late is not covered. These let us distinguish between error types, including situations when the graph missed the page, when retrieval picked the wrong target, and when the route itself was noisy.

\subsection{Strict Judge and Calibration}

Primary navigation results use human judgments. Strict success requires the reference destination, allowing cosmetic differences on the same functional page; near misses fail. GPT-4o~\citep{hurst2024gpt4o} cross-checks final and reference screenshots on a 1--5 scale (4--5 indicate success); on a 530-pair calibration set it agrees with human judgments on 97.5\% of binary labels (Appendix~\ref{app:judge-calibration}).

\section{Experiments}
\label{sec:experiments}

\subsection{Setup}
\label{sec:exp-setup}

\paragraph{Backbones and protocols.}
We wrap four visual GUI actors without fine-tuning: UI-TARS-1.5-7B \citep[UI-TARS;][]{qin2025uitars} and Qwen-VL-Max~\citep[Qwen;][]{bai2025qwen25vl}\footnote{Queried through the DashScope API model alias \texttt{qwen-vl-max-latest} (April--May 2026).} as headline backbones, plus AgentCPM-GUI~\citep{zhang2025agentcpmgui} (8B open-weight MiniCPM-V-based) and MAI-UI~\citep{zhou2025maiui} as cross-family checks (Table~\ref{tab:main-result-cost}). Benchmarks used in our experiments are \benchmark\ (Sec.~\ref{sec:destbench}), DroidTask \citep{wen2024autodroid}, and SPA-Bench \citep{chen2024spabench}. For DroidTask/SPA-Bench we redefine success as reaching the pre-action destination (not the final action), reusing the benchmark-native judge (Appendix~\ref{app:reimpl}) and reporting final numbers using human-annotated results. Only within-column \mbox{\texttt{no-memory $\rightarrow$}} \texttt{+\method} deltas are compared. Table~\ref{tab:graph-stats-coverage} gives per-app graph sizes and target coverage (exploration effort: Appendix~\ref{app:local-bfs}). Online navigation uses the canonical configuration: dual-field retrieval, top-1 selection, waypoint memory (Appendix~\ref{app:protocol}).

\subsection{Main Result: Cross-Backbone, Cross-Benchmark}
\label{sec:exp-main}

\begin{figure*}[!t]
\centering
\definecolor{tagblue}{HTML}{4A78C8}
\definecolor{tagbluelight}{HTML}{BFD1ED}
\definecolor{tagpurple}{HTML}{7E68B5}
\definecolor{tagpurplelight}{HTML}{CFC2E2}
\begin{tikzpicture}
\begin{axis}[
    ybar=0.6pt,
    width=0.82\textwidth,
    height=2.7cm,
    bar width=8pt,
    enlarge x limits=0.08,
    ymin=0, ymax=100,
    ytick={0,20,40,60,80,100},
    ylabel={SR (\%)},
    ylabel style={font=\scriptsize, yshift=-4pt},
    symbolic x coords={Amazon, Calendar, Clock, Maps, Settings, YouTube},
    xtick=data,
    xticklabel style={font=\scriptsize, yshift=-1pt},
    yticklabel style={font=\scriptsize},
    ymajorgrids=true,
    grid style={dashed, line width=0.25pt, draw=gray!40},
    axis line style={draw=gray!60, line width=0.4pt},
    tick style={draw=gray!60, line width=0.3pt},
    legend style={
        font=\scriptsize,
        at={(0.5, 1.03)},
        anchor=south,
        legend columns=4,
        draw=none,
        fill=white,
        fill opacity=0.8,
        text opacity=1,
        column sep=4pt,
        /tikz/every even column/.append style={column sep=6pt},
        cells={anchor=west},
    },
    legend image code/.code={\draw [#1] (0cm,-0.07cm) rectangle (0.22cm, 0.15cm);},
    clip=false,
]
\addplot[fill=tagbluelight, draw=tagblue, line width=0.35pt]
    coordinates {(Amazon, 53) (Calendar, 50) (Clock, 11) (Maps, 50) (Settings, 89) (YouTube, 60)};
\addplot[fill=tagblue, draw=tagblue!75!black, line width=0.35pt]
    coordinates {(Amazon, 59) (Calendar, 40) (Clock, 33) (Maps, 63) (Settings, 78) (YouTube, 67)};
\addplot[fill=tagpurplelight, draw=tagpurple, line width=0.35pt]
    coordinates {(Amazon, 47) (Calendar, 70) (Clock, 44) (Maps, 50) (Settings, 78) (YouTube, 80)};
\addplot[fill=tagpurple, draw=tagpurple!75!black, line width=0.35pt]
    coordinates {(Amazon, 65) (Calendar, 70) (Clock, 78) (Maps, 71) (Settings, 78) (YouTube, 73)};

\legend{UI-TARS-1.5-7B, \,+\,\method, Qwen-VL-Max, \,+\,\method}
\end{axis}
\end{tikzpicture}
\caption{\textbf{Per-app navigation success rate with and without \method.} Gains are largest when navigation is the bottleneck (Clock, Amazon, Google Maps) and smaller when the baseline is already strong or execution dominates (Settings, Calendar).}
\label{fig:per-app-delta}
\end{figure*}
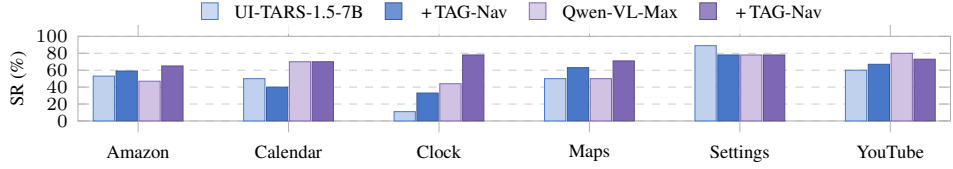

\looseness=-1 Table~\ref{tab:cross-backbone} reports the headline result: \method improves over no-memory in every aggregate cell. These are navigation gains, not end-to-end task-completion gains. Graph coverage is partial (Table~\ref{tab:graph-stats-coverage}), and a backbone can still reach an uncovered destination from its own app knowledge, so successes are not bounded by coverage. The aggregate masks per-app heterogeneity (Figure~\ref{fig:per-app-delta}): apps whose residual difficulty is execution rather than navigation (Calendar, Settings) show zero or negative per-cell deltas, so \method\ is a navigation-memory aid rather than a uniform improvement; see \S\ref{sec:limitations}. Settings already has a strong no-memory baseline, and Calendar failures often occur at terminal date/time manipulation after reaching the relevant page. Tables~\ref{tab:cross-backbone} and~\ref{tab:original-paired-results} report paired tests (per-app uncertainty: Appendix~\ref{app:historical-uncertainty}); Table~\ref{tab:per-app-droidtask} gives per-app DroidTask-Nav and SPA-Bench-Nav results.

\begin{table}[!htbp]
\centering
\scriptsize
\setlength{\tabcolsep}{1.4pt}
\renewcommand{\arraystretch}{0.92}
\begin{tabularx}{\columnwidth}{@{}>{\raggedright\arraybackslash}p{0.24\columnwidth}>{\centering\arraybackslash}p{0.24\columnwidth}>{\centering\arraybackslash}p{0.24\columnwidth}>{\centering\arraybackslash}p{0.24\columnwidth}@{}}
\toprule
\textbf{Backbone} & \textbf{DestBench} & \textbf{DroidTask-Nav} & \textbf{SPA-Bench-Nav} \\
 & SR (\%) & SR (\%) & SR (\%) \\
\midrule
UI-TARS-1.5-7B & 52.4 $\rightarrow$ \textbf{58.3} & 63.3 $\rightarrow$ \textbf{71.3} & 66.7 $\rightarrow$ \textbf{76.7} \\
\quad $p$ & 0.383 & 0.073 & 0.508 \\
\addlinespace[0.2em]
Qwen-VL-Max & 59.5 $\rightarrow$ \textbf{71.4} & 66.0 $\rightarrow$ \textbf{70.7}$^{\S}$ & 76.7 $\rightarrow$ \textbf{93.3} \\
\quad $p$ & 0.076 & 0.265 & 0.125 \\
\midrule
Avg.\ gain & +8.9 & +6.4 & +13.3 \\
\bottomrule
\end{tabularx}
\caption{\textbf{Cross-backbone, cross-benchmark navigation.} Cells report \texttt{no-memory $\rightarrow$ +\method} in success rate (SR). DroidTask and SPA-Bench are converted to page-reaching tasks; only within-column deltas are comparable. All cells use human-annotated strict success. $p$: exact McNemar test; 95\% confidence intervals (CIs) are in Appendix~\ref{app:historical-uncertainty}. $^{\S}$The no-memory arm resized screenshots to a 1,280-pixel longest side; the +\method\ arm used full resolution.}
\label{tab:cross-backbone}
\end{table}

\begin{table}[!htbp]
\centering
\scriptsize
\setlength{\tabcolsep}{3pt}
\begin{tabular}{@{}lrrrr@{}}
\toprule
\textbf{Graph} & \textbf{Success} & \textbf{Gain (pp)} & \textbf{95\% CI (pp)} & \textbf{$p$} \\
\midrule
\multicolumn{5}{@{}l}{\textit{UI-TARS-1.5-7B}} \\
No memory & 44/84 & -- & -- & -- \\
Random & 54/84 & +11.9 & [0.0, 23.8] & 0.087 \\
Auto-BFS & 49/84 & +6.0 & [$-4.8$, 16.7] & 0.383 \\
Manual & 59/84 & +17.9 & [8.3, 27.4] & 0.0007 \\
\midrule
\multicolumn{5}{@{}l}{\textit{Qwen-VL-Max}} \\
No memory & 50/84 & -- & -- & -- \\
Random & 54/84 & +4.8 & [$-7.1$, 16.7] & 0.572 \\
Auto-BFS & 60/84 & +11.9 & [0.0, 23.8] & 0.076 \\
Manual & 56/84 & +7.1 & [$-4.8$, 19.0] & 0.345 \\
\bottomrule
\end{tabular}
\caption{\textbf{Paired graph-condition comparisons on DestBench.} Success is the human-judged destination match count. Gains and 95\% CIs are in percentage points (pp) relative to no memory for the same backbone; $p$: exact McNemar test (Appendix~\ref{app:historical-uncertainty}). Some random- and manual-arm cells used the same app's memory from a different graph condition (Appendix~\ref{app:graph-arm-provenance}); without those cells, every gain remains positive.}
\label{tab:original-paired-results}
\end{table}

\begin{table}[!t]
\centering
\scriptsize
\setlength{\tabcolsep}{1.5pt}
\renewcommand{\arraystretch}{0.95}
\begin{tabularx}{\columnwidth}{@{}>{\raggedright\arraybackslash}X>{\centering\arraybackslash}p{0.15\columnwidth}>{\centering\arraybackslash}p{0.15\columnwidth}>{\centering\arraybackslash}p{0.15\columnwidth}>{\centering\arraybackslash}p{0.15\columnwidth}@{}}
\toprule
 & \multicolumn{2}{c}{\textbf{No memory}} & \multicolumn{2}{c}{\textbf{+ \method}} \\
\cmidrule(lr){2-3}\cmidrule(lr){4-5}
\textbf{App} & \shortstack{UI-TARS-\\1.5-7B} & \shortstack{Qwen-VL-\\Max} & \shortstack{UI-TARS-\\1.5-7B} & \shortstack{Qwen-VL-\\Max} \\
\midrule
\multicolumn{5}{@{}l}{\textit{DroidTask-Nav (12 apps, 150 tasks)}} \\
AppLauncher   & 2/5   & 4/5   & 5/5   & 2/5   \\
Calendar      & 11/17 & 9/17  & 12/17 & 12/17 \\
Camera        & 5/15  & 5/15  & 4/15  & 5/15  \\
Clock         & 7/12  & 9/12  & 11/12 & 11/12 \\
Contacts      & 12/14 & 14/14 & 13/14 & 13/14 \\
Dialer        & 7/15  & 12/15 & 6/15  & 12/15 \\
FileManager   & 6/16  & 10/16 & 14/16 & 13/16 \\
Gallery       & 8/9   & 8/9   & 9/9   & 8/9   \\
Messenger     & 13/15 & 8/15  & 11/15 & 11/15 \\
MusicPlayer   & 7/9   & 7/9   & 7/9   & 7/9   \\
Notes         & 9/14  & 5/14  & 6/14  & 4/14  \\
VoiceRecorder & 8/9   & 8/9   & 9/9   & 8/9   \\
\midrule
\textbf{Total} & 95/150 & 99/150 & 107/150 & 106/150 \\
\textbf{SR (\%)} & 63.3 & 66.0 & 71.3 & 70.7 \\
\midrule
\multicolumn{5}{@{}l}{\textit{SPA-Bench-Nav (7 apps, 30 tasks)}} \\
Airbnb      & 1/3 & 0/3 & 1/3 & 2/3 \\
Amazon      & 2/3 & 3/3 & 3/3 & 3/3 \\
Calendar    & 3/3 & 3/3 & 2/3 & 3/3 \\
Chrome      & 2/3 & 3/3 & 3/3 & 3/3 \\
Clock       & 6/6 & 3/6 & 6/6 & 5/6 \\
Google Maps & 3/6 & 5/6 & 5/6 & 6/6 \\
YouTube     & 3/6 & 6/6 & 3/6 & 6/6 \\
\midrule
\textbf{Total} & 20/30 & 23/30 & 23/30 & 28/30 \\
\textbf{SR (\%)} & 66.7 & 76.7 & 76.7 & 93.3 \\
\bottomrule
\end{tabularx}
\caption{\textbf{DroidTask-Nav and SPA-Bench-Nav per-app drilldowns.} Tasks converted from task-completion to navigation-only goals. Human-annotated strict destination success for both benchmarks.}
\label{tab:per-app-droidtask}\label{tab:per-app-spa}
\end{table}

\subsection{Graph Quality: Controlled Decomposition}
\label{sec:exp-graph-quality}

Table~\ref{tab:graph-quality} decomposes the automated-to-manual gap on Qwen. Row (d) keeps the automated waypoint pipeline but replaces the retrieved target with the manual destination (isolating retrieval); it does not improve over \method\ (70.2 vs.\ 71.4); per-app drilldown in Appendix~\ref{app:per-app-graph-quality}. Row (f) (manual exploration) sits \emph{below} Auto-BFS (66.7 vs.\ 71.4) because manual graphs trade coverage for cleanliness; it is a ``clean routes, narrow coverage'' condition rather than an upper bound. In a separate observational analysis of 313 archived memory-augmented runs judged by GPT-4o, success falls from 58\% for routes with 1--2 waypoints to 50\% for 3--4 and 38\% for 5 or more (correlation analysis in Appendix~\ref{app:route-cleanliness}). Appendix~\ref{app:response-period} varies which record fields are injected.

\begin{table}[!htbp]
\centering
\scriptsize
\setlength{\tabcolsep}{4pt}
\renewcommand{\arraystretch}{1.1}
\begin{tabularx}{\columnwidth}{@{}>{\raggedright\arraybackslash}X>{\centering\arraybackslash}p{0.13\columnwidth}>{\centering\arraybackslash}p{0.13\columnwidth}>{\centering\arraybackslash}p{0.13\columnwidth}@{}}
\toprule
\textbf{Condition} & \textbf{Targets} & \textbf{Routes} & \textbf{SR (\%)} \\
\midrule
(a) No memory & -- & -- & 59.5 \\
(b) Random exploration & auto & noisy & 64.3 \\
\midrule
\textbf{(c) \method (Auto-BFS)} & auto & auto & \textbf{71.4} \\
\midrule
(d) Manual targets + auto routes & manual & auto & 70.2 \\
(e) Auto targets + manual routes & auto & manual & \textbf{79.8} \\
\midrule
(f) Manual exploration & manual & manual & 66.7 \\
\bottomrule
\end{tabularx}
\caption{\textbf{Graph-quality controlled decomposition on \benchmark\ (Qwen-VL-Max).} Human-annotated strict success on the 84-task core. Row (d) ran at full input resolution on a separate host. Rows (d) and (e) replace one memory field at a time with manual ground truth. Rows (b) and (f) include app cells that used the same app's memory from a different graph condition (Appendix~\ref{app:graph-arm-provenance}). The UI-TARS counterpart is in Appendix~\ref{app:per-app-graph-quality}.}
\label{tab:graph-quality}
\end{table}

\subsection{Comparison to Prior Memory Systems}
\label{sec:exp-vs-prior}

Table~\ref{tab:main-result-cost} compares \method against prior memory-augmented GUI agents as a memory-format comparison on our task set and human-annotated judge, not as a claim of outperforming PG-Agent or GUI-Explorer in their released forms. PG-Agent, which needs human demonstrations, trails no-memory UI-TARS, and GUI-Explorer (GPT-4o actor) falls between UI-TARS without and with \method, while \method adds unsupervised exploration plus one retrieval per task and improves every wrapped backbone (adaptation rules and actors in App.~\ref{app:reimpl}). These rows isolate the memory abstraction (custom validators and planners omitted) and do not reflect each system's published end-to-end performance.

\begin{table}[!htbp]
\centering
\scriptsize
\setlength{\tabcolsep}{4pt}
\renewcommand{\arraystretch}{1.05}
\begin{tabularx}{\columnwidth}{@{}Xrr@{}}
\toprule
\textbf{System} & \textbf{Success} & \textbf{SR (\%)} \\
\midrule
\multicolumn{3}{@{}l}{\textit{Prior memory-format adaptations}} \\
PG-Agent$^{\dagger}$ & 36/84 & 42.9 \\
GUI-Explorer & 47/84 & 56.0 \\
\midrule
\multicolumn{3}{@{}l}{\textit{Backbone agents without / with \method}} \\
AgentCPM-GUI & 28/84 & 33.3 \\
\quad + \method & 33/84 & 39.3 \scriptsize{(+6.0)} \\
\addlinespace[0.3em]
UI-TARS-1.5-7B & 44/84 & 52.4 \\
\quad + \method & 49/84 & 58.3 \scriptsize{(+5.9)} \\
\addlinespace[0.3em]
MAI-UI & 43/78 & 55.1 \\
\quad + \method & 47/79 & 59.5 \scriptsize{(+4.4)} \\
\addlinespace[0.3em]
Qwen-VL-Max & 50/84 & 59.5 \\
\quad + \method & 60/84 & 71.4 \scriptsize{(+11.9)} \\
\bottomrule
\end{tabularx}
\caption{\textbf{Prior memory formats and wrapped backbones on \benchmark.} \method adds one-time unsupervised exploration and one retrieval call per task. All rows use human-annotated strict success.
$^{\dagger}$PG-Agent additionally requires human-labeled task demonstrations.}
\label{tab:main-result-cost}
\end{table}

\subsection{Transfer to Additional Apps}
\label{sec:exp-additional}

To test whether the benefit extends beyond the six core apps, we ran UI-TARS-1.5-7B with and without Auto-BFS memory on 68 tasks from five further apps, labeled with the same human strict rule. Memory raises success from 28/68 to 35/68 (+10.3 pp; 11 tasks improved, 4 regressed; exact McNemar $p=0.12$), with gains on four of the five apps and a loss on Instagram (Table~\ref{tab:five-app-cohort}). The direction matches the core results, though the gain is not significant in this small cohort.

\begin{table}[t]
\centering
\scriptsize
\begin{tabular}{lrr}
\toprule
\textbf{App} & \textbf{No memory} & \textbf{+ \method} \\
\midrule
Airbnb & 7/13 & 10/13 \\
Chrome & 4/13 & 6/13 \\
eBay & 11/16 & 13/16 \\
Instagram & 4/13 & 2/13 \\
TikTok & 2/13 & 4/13 \\
\midrule
Total & 28/68 (41.2\%) & 35/68 (51.5\%) \\
\bottomrule
\end{tabular}
\caption{\textbf{Transfer to five additional apps.} UI-TARS-1.5-7B, human-judged strict destination match; separate from the 84-task core.}
\label{tab:five-app-cohort}
\end{table}

\subsection{Failure Taxonomy}
\label{sec:exp-failures}

Of 48 labeled open-loop \method failures, 22 are wrong retrievals (A), 15 near misses (B), and 11 execution failures (C); A and B together account for $\sim$77\% and are addressable to the memory pipeline, so destination retrieval and target selection remain fragile (per-class notes in \S\ref{sec:limitations} and App.~\ref{app:failure-trajectories}).

\section{Conclusion}
\label{sec:conclusion}

\method introduces a destination-memory module that wraps existing visual GUI agents without retraining, converting task-agnostic offline exploration into compact target and route records injected at inference time.
Across two backbones and three benchmarks, this memory improves aggregate navigation success, and graph-quality controls associate success more with route cleanliness than with graph size.
This makes app-specific memory a measurable destination map rather than a larger exploration trace.

\clearpage

\section{Limitations}
\label{sec:limitations}

Current evidence is strongest for Android app-navigation tasks and a Qwen-VL-Max backbone.
Auto-BFS graph memory still trails clean/manual memory in some conditions.
Graph memory depends on app version, account/login state, UI drift, and exploration coverage.
Apps with sufficiently large navigation-affecting UI updates require re-exploration before previously collected destination memory can be trusted; in a small two-app probe, full-record memory gave no benefit over no memory before or after updates (Table~\ref{tab:drift-probe}).
\method is also not plug-and-play with arbitrary task-completion benchmarks because it requires page-reaching targets, safe stopping points, and graph-aware diagnostics.
Manual/clean graph results should be interpreted as a control rather than the deployment assumption.
Comparisons to external benchmarks require careful protocol matching.
\method is most useful when the destination is graph-covered, the route is short and clean, and the backbone can reliably execute visible actions; coverage gaps, app-state drift, and sibling-page retrieval errors are the dominant remaining failure modes.
The three behavioral failure classes documented in Table~\ref{tab:failure-modes} (\S\ref{sec:exp-failures}) correspond to these limits: \emph{wrong retrieval} (Class A) reflects coverage and embedding limits, \emph{near miss} (Class B) reflects last-step disambiguation limits that sharper target selection and cleaner routes may address, and \emph{execution failure} (Class C) reflects step-budget exhaustion and open-loop route-following limits that per-step memory refresh may address.

\paragraph{Per-app heterogeneity.}
Gains are not uniform across apps (Figure~\ref{fig:per-app-delta}): Calendar and Settings, where the residual difficulty is execution rather than navigation, show smaller or negative deltas (\S\ref{sec:exp-main}). We do not claim \method improves every (app, backbone) cell.

\section{Ethical Considerations}
\label{sec:ethical-considerations}

\noindent\textbf{Authorized test setting and privacy.}
\method exercises third-party mobile applications, so all exploration and evaluation used dedicated test devices, disposable accounts, and locally collected test data, not personal user accounts.
\benchmark tasks are derived from existing benchmark trajectories and our disposable-account collection; released artifacts use scrubbed or generic targets, and we do not distribute user-identifiable content.
Screenshots captured during exploration may incidentally contain account-specific content; explored graphs are intended for the same app version and account state in which they were collected, and we recommend that downstream users re-collect graphs under their own accounts rather than reusing ours when account state matters.
Downstream use should respect app-level terms of service and obtain authorization where required.

\noindent\textbf{Dual use and release.}
Autonomous mobile GUI agents can in principle be repurposed for spam, automated account creation, or unauthorized actions on user accounts.
\method is a navigation-memory module for existing backbones rather than a new execution capability: it improves where the agent goes, not what novel actions it can perform.
\benchmark is intentionally scoped to page reaching and excludes the irreversible final actions most relevant to abuse scenarios such as purchasing, sending, deleting, or changing sensitive settings.
All backbone models are used under their respective open-weight or API licenses.
We plan to release code, exploration scripts, and the evaluation protocol under terms limiting use to authorized accounts and test environments; app graphs and trajectories will be shared only where the underlying screenshots can be released.

\noindent\textbf{Compute and environmental impact.}
Offline exploration and online evaluation run on commodity Android emulators; open-weight backbones run on a local GPU host, while Qwen models and GPT-4o are queried through the DashScope and OpenAI APIs.
We do not train a new GUI backbone in this work, so compute usage is modest relative to model-training contributions, and the offline exploration cost amortizes across all later navigation tasks for the same app version.
At inference time, \method\ adds one retrieval call per task to the backbone's own calls; graph-construction counters are reported in Appendix~\ref{app:cost}.

\noindent\textbf{Data minimization and annotator labor.}
\benchmark\ task descriptions and final-screenshot targets were annotated by the authors.
Strict-judge calibration used the same pool to label 530 (task, final-screenshot) pairs, with labelers shown only the two screens and the user instruction.

\clearpage

\bibliography{references/custom}

\clearpage
\onecolumn

\appendix

\section*{Appendix}
\label{sec:appendix}

\vspace{-0.75em}
\subsection*{Table of Contents}
\vspace{-0.25em}
\begin{center}
\small
\renewcommand{\arraystretch}{1.0}
\setlength{\tabcolsep}{6pt}
\begin{tabularx}{\textwidth}{@{}p{0.045\textwidth}X p{0.07\textwidth}@{}}
\toprule
 & \textbf{Section} & \textbf{Page} \\
\midrule
\textbf{A} & \textbf{\benchmark\ Construction}                                          & \pageref{app:destbench-construction}\\
 & A.1\quad App selection rationale and full app list                                    & \pageref{app:app-selection}\\
 & A.2\quad Task-collection pipeline                                                     & \pageref{app:task-collection}\\
 & A.3\quad Destination annotation protocol                                              & \pageref{app:annotation}\\
 & A.4\quad Representative tasks                                                         & \pageref{app:sample-tasks}\\
 & A.5\quad GPT-4o judge calibration                                                     & \pageref{app:judge-calibration}\\
 & A.6\quad \benchmark\ app counts                                                       & \pageref{app:destbench-app-counts}\\
\addlinespace[0.3em]
\textbf{B} & \textbf{\method\ Implementation Details}                                   & \pageref{app:implementation}\\
 & B.1\quad Hypernode matching                                                           & \pageref{app:hypernode-match}\\
 & B.2\quad Local-BFS walk policy                                                    & \pageref{app:local-bfs}\\
 & B.3\quad Memory record schema                                                         & \pageref{app:record-schema}\\
 & B.4\quad Backbone configurations                                                      & \pageref{app:backbone-configs}\\
 & B.5\quad Protocol details                                                             & \pageref{app:protocol}\\
\addlinespace[0.3em]
\textbf{C} & \textbf{Prompts}                                                           & \pageref{app:prompts}\\
 & C.1\quad Metadata extraction prompt (page summary, direct intent, route distillation) & \pageref{app:prompt-summary}\\
 & C.2\quad Target selector prompt                                                       & \pageref{app:prompt-selector}\\
 & C.3\quad Destination-memory injection template                                        & \pageref{app:prompt-injection}\\
 & C.4\quad GPT-4o judge prompt                                                          & \pageref{app:prompt-judge}\\
\addlinespace[0.3em]
\textbf{D} & \textbf{Additional Quantitative Results}                                   & \pageref{app:additional-results}\\
 & D.1\quad Per-app, per-backbone, per-benchmark breakdown & \pageref{app:per-app}\\
 & D.2\quad Per-app uncertainty & \pageref{app:historical-uncertainty}\\
 & D.3\quad Graph-arm memory provenance & \pageref{app:graph-arm-provenance}\\
 & D.4\quad Per-app graph-quality breakdown & \pageref{app:per-app-graph-quality}\\
 & D.5\quad Route cleanliness & \pageref{app:route-cleanliness}\\
 & D.6\quad Memory format and retrieval-depth ablation & \pageref{app:memory-format}\\
 & D.7\quad Cross-family and prior-memory outcomes & \pageref{app:cross-family-outcomes}\\
 & D.8\quad Cost analysis & \pageref{app:cost}\\
 & D.9\quad Record-content ablation & \pageref{app:response-period}\\
 & D.10\quad Informed no-memory baseline & \pageref{app:informed-baseline}\\
\addlinespace[0.3em]
\textbf{E} & \textbf{Format-Adapted Reimplementation Protocol}                          & \pageref{app:reimpl}\\
 & E.1\quad Per-system adaptation rules                                                  & \pageref{app:reimpl-rules}\\
\addlinespace[0.3em]
\textbf{F} & \textbf{Qualitative Examples}                                              & \pageref{app:qualitative}\\
 & F.1\quad Failure trajectories per class                                               & \pageref{app:failure-trajectories}\\
 & F.2\quad Success trajectories                                                         & \pageref{app:success-trajectories}\\
 & F.3\quad Raw destination-record examples                                              & \pageref{app:record-examples}\\
\addlinespace[0.3em]
\textbf{G} & \textbf{Detailed Related-Work Positioning}                                 & \pageref{app:related-positioning}\\
\addlinespace[0.3em]
\textbf{H} & \textbf{Limitations, Robustness, and Ethics}                               & \pageref{app:limits}\\
 & H.1\quad Dynamic-screen handling                                                      & \pageref{app:dynamic-screens}\\
 & H.2\quad Cross-app-version robustness                                                 & \pageref{app:version-robustness}\\
 & H.3\quad Multi-path and ambiguous-target handling                                     & \pageref{app:multipath}\\
 & H.4\quad Privacy and sensitive-screen handling                                        & \pageref{app:privacy}\\
 & H.5\quad Use of AI in paper preparation                                               & \pageref{app:ai-paper-usage}\\
 & H.6\quad Models used in the \method\ pipeline                                         & \pageref{app:llm-usage}\\
 & H.7\quad Dynamic-memory variants                                                      & \pageref{app:dynamic-variants}\\
\addlinespace[0.3em]
\bottomrule
\end{tabularx}
\end{center}

\clearpage
\twocolumn
\small
\setlength{\textfloatsep}{6pt plus 1pt minus 2pt}
\setlength{\floatsep}{6pt plus 1pt minus 2pt}
\setlength{\intextsep}{6pt plus 1pt minus 2pt}

\section{\benchmark\ Construction}
\label{app:destbench-construction}

This section documents the construction of \benchmark, from app selection through annotation through automated judging.

\subsection{App selection rationale and full app list}
\label{app:app-selection}
We selected 20 Android apps spanning nine product categories, with a core cohort of 6 apps used for the controlled graph-condition study (no-memory, random, Auto-BFS, clean manual). Apps were chosen to span (i) shallow vs deep navigation depth, (ii) static vs dynamic content, (iii) presence vs absence of account-gated screens, and (iv) presence of modals, drawers, and bottom-sheet UI patterns. Table~\ref{tab:destbench-apps} lists all apps with their category and core-cohort membership.

\begin{table}[H]
\centering
\footnotesize
\setlength{\tabcolsep}{4pt}
\renewcommand{\arraystretch}{1.05}
\begin{tabularx}{\columnwidth}{@{}lXc@{}}
\toprule
\textbf{App} & \textbf{Category} & \textbf{Core 6} \\
\midrule
Amazon                & Shopping               & $\bullet$ \\
Calendar              & Productivity           & $\bullet$ \\
Clock                 & Utility                & $\bullet$ \\
Google Maps           & Travel                 & $\bullet$ \\
Settings              & Utility                & $\bullet$ \\
YouTube               & Entertainment          & $\bullet$ \\
\addlinespace[0.2em]
Airbnb                & Travel                 & \\
Chrome                & Productivity           & \\
eBay                  & Shopping               & \\
Instagram             & Social Media           & \\
LinkedIn              & Social Media           & \\
Spotify               & Entertainment          & \\
Starbucks             & Food \& Drink          & \\
Target                & Shopping               & \\
TikTok                & Social Media           & \\
Uber                  & Travel                 & \\
Uber Eats             & Food \& Drink          & \\
Walmart               & Shopping               & \\
WhatsApp              & Communication          & \\
Yelp                  & Local Services         & \\
\bottomrule
\end{tabularx}
\caption{\textbf{Full \benchmark app inventory.} 20 Android apps spanning Shopping, Productivity, Utility, Travel, Entertainment, Social Media, Food \& Drink, Communication, and Local Services categories. The 6-app core cohort is the subset used for the controlled graph-condition study (no-memory, random, automated Local-BFS, clean manual).}
\label{tab:destbench-apps}
\end{table}

\subsection{Task-collection pipeline}
\label{app:task-collection}
Tasks are derived from real user-trajectory recordings collected in our benchmark substrate. Each recording is segmented at the pre-action destination boundary (the last screen before a task-specific irreversible operation, defined in \S\ref{sec:destbench}) and labeled with a natural-language instruction by paraphrasing the original goal. Instructions are filtered to remove user-identifying content and to disambiguate destinations that admit multiple visually-distinct correct pages.

\subsection{Destination annotation protocol}
\label{app:annotation}
Each task instance is annotated with a target screenshot $x_{\mathrm{target}}$ that visually identifies the pre-action destination, plus a dynamic-state tolerance mask that lists which UI regions (e.g., timestamps, scroll positions, advertisement slots) are ignored by the judge. The primary curator annotated the full core cohort; a second author audited a random 20\% sample of the annotated targets, and the small number of disagreements (predominantly cases where multiple visually-distinct pages plausibly satisfied the instruction) were resolved jointly against the original trajectory recording. We report this rather than a full second-pass inter-annotator agreement because the dominant source of annotation noise in our pipeline is target ambiguity, not labeler disagreement on unambiguous cases, and the strict-judge calibration in App.~\ref{app:judge-calibration} provides a downstream check on residual labeling noise. The primary metric requires the reference destination; a near miss is a failure.

\subsection{Representative tasks}
\label{app:sample-tasks}
Table~\ref{tab:sample-tasks} gives six representative tasks across six apps, showing the instruction, target page, and the stopping boundary that excludes downstream execution from scoring.

\begin{table*}[!tb]
\centering
\footnotesize
\setlength{\tabcolsep}{4pt}
\renewcommand{\arraystretch}{1.15}
\begin{tabularx}{\linewidth}{@{}p{0.10\linewidth}p{0.34\linewidth}p{0.27\linewidth}p{0.20\linewidth}@{}}
\toprule
\textbf{App} & \textbf{User instruction $q$} & \textbf{Target page (destination)} & \textbf{Stopping boundary} \\
\midrule
Amazon & ``View past orders.''                                & ``Your Orders'' history page                & Before opening any order detail \\
Calendar & ``Change the default reminder time for events to 15 minutes.'' & ``Default notification time'' settings row & Before tapping the time picker \\
TikTok & ``Open the privacy settings for who can comment on my videos.'' & ``Who can comment'' configuration page & Before selecting a permission \\
Chrome & ``Clear browsing history from the last hour.''       & ``Clear browsing data'' time-range dialog   & Before tapping ``Clear data'' \\
Google Maps & ``Switch the map style to satellite.''          & Map-style selector with satellite option    & Before confirming the style change \\
WhatsApp & ``Turn off read receipts.''                        & ``Read receipts'' toggle row                & Before flipping the toggle \\
\bottomrule
\end{tabularx}
\caption{\textbf{Representative \benchmark task examples.} Each task specifies a natural-language instruction $q$, a pre-action destination screen the agent must reach, and a safe stopping boundary that excludes the irreversible final operation. Success is judged by visual equivalence between the agent's final screenshot and the destination annotation; the stopping-boundary action is never executed in scoring.}
\label{tab:sample-tasks}
\end{table*}

\subsection{GPT-4o judge calibration}
\label{app:judge-calibration}
The judge prompt (Appendix~\ref{app:prompt-judge}) is calibrated against a single-curator gold set of 530 (task, final-screenshot) pairs. Table~\ref{tab:judge-calibration} reports binary agreement, false-pass/false-fail breakdown, and ordinal $\kappa$. The residual 13 disagreements are primarily false-pass, and concentrate on near-miss cases (sibling pages, stopping one page early). The ordinal Cohen's $\kappa$ of 0.95 indicates substantial agreement beyond chance.

\begin{table}[H]
\centering
\footnotesize
\setlength{\tabcolsep}{6pt}
\renewcommand{\arraystretch}{1.1}
\begin{tabular}{lr}
\toprule
\textbf{Metric} & \textbf{Value} \\
\midrule
Total calibration tasks     & 530 \\
Binary agreement            & 517 / 530 (97.5\%) \\
\midrule
False pass  & 10 \\
False fail  & 3 \\
\midrule
Cohen's $\kappa$ (binary) & 0.95 \\
\bottomrule
\end{tabular}
\caption{\textbf{GPT-4o judge calibration against human annotators.} Calibration set: 530 (task, agent-final-screenshot) pairs spanning all 20 apps. Binary agreement represents matching judge and human results. False pass represents a reported success from the judge, and failure from the human annotator, while false fail represents a reported failure from the judge, and success from the human annotator. Binary success label uses scores 4--5 = success.}
\label{tab:judge-calibration}
\end{table}

\subsection{\benchmark\ app counts}
\label{app:destbench-app-counts}
Table~\ref{tab:destbench-task-distribution} (rendered inline in \S\ref{sec:destbench}) shows per-app task counts in the core cohort. The 6-app core cohort spans Google Maps, Amazon, YouTube, Settings, Clock, and Calendar, totaling 84 tasks.

\section{\method\ Implementation Details}
\label{app:implementation}

\subsection{Hypernode matching}
\label{app:hypernode-match}
Hypernode matching compares a new screen against each stored member screen of a hypernode on two signals: visual similarity, as the cosine similarity of CLIP ViT-B/32 screenshot embeddings~\citep{radford2021clip}, and layout similarity of the Android XML view hierarchy, using a normalized tree-edit distance and \texttt{text-embedding-v4} embeddings of the page layout. A screen joins the hypernode when both signals are within thresholds tuned during graph collection.

\subsection{Local-BFS walk policy}
\label{app:local-bfs}
Exploration runs up to 1,000 walks from the app root, each with at most five forward steps. With probability 0.7, a walk expands locally: it tries untested actions on the current screen and verifies each return, for up to 15 successful undo-and-check cycles; otherwise it explores randomly. Actions already tried on the current screen or its matched siblings are skipped, and the rest are ordered by type (image elements first, then scroll, swipe, and card actions, then the rest), with random tie-breaking. Table~\ref{tab:cost-analysis} reports the resulting walk and action counts per app.

\subsection{Memory record schema}
\label{app:record-schema}
Each destination record $r_{h,n}$ is a tuple $(h, n, d, I, W, H, z^d, z^I)$ where $h$ is the hypernode id, $n$ is the member-node id, $d$ is a natural-language page summary, $I$ is a list of direct user-intents the page satisfies, $W$ is a deduplicated waypoint sequence from the app root, $H$ is a list of per-transition disambiguation hints, and $z^d, z^I$ are DashScope \texttt{text-embedding-v4} embeddings~\citep{zhang2025qwen3embedding} of $d$ and the concatenation of $I$. Records are stored as JSONL with one record per line for fast top-$k$ retrieval at inference.

\subsection{Backbone configurations}
\label{app:backbone-configs}
Table~\ref{tab:backbone-configs} lists the shared prompting and decoding settings used for both backbones.

\begin{table}[H]
\centering
\footnotesize
\setlength{\tabcolsep}{4pt}
\renewcommand{\arraystretch}{1.1}
\begin{tabularx}{\columnwidth}{@{}lXXX@{}}
\toprule
\textbf{Setting} & \textbf{UI-TARS-1.5-7B} & \textbf{Qwen-VL-Max} \\
\midrule
Image resolution        & native (1080p portrait) & native \\
Max input tokens        & 8192                    & API-controlled \\
Max output tokens       & 512                     & API-controlled \\
Temperature             & 0.0                     & 0.0  \\
Top-$p$                 & 1.0                     & 1.0 \\
Action space            & mobile-ops              & mobile-ops \\
History window (steps)  & 3                       & 3 \\
Step budget $T$         & 15                      & 15 \\
\bottomrule
\end{tabularx}
\caption{\textbf{Backbone configurations.} Same prompting and decoding hyperparameters across both backbones; only the model weights and tokenizer differ. \texttt{mobile-ops} is the shared action vocabulary (\texttt{click}, \texttt{type}, \texttt{scroll}, \texttt{back}, \texttt{home}, \texttt{wait}, \texttt{finished}).}
\label{tab:backbone-configs}
\end{table}

\subsection{Protocol details}
\label{app:protocol}
Device configuration (Android emulator on Android 16, API level 36.1, ABI arm64-v8a), action space, retrieval top-$k$, step budgets, and seed/temperature settings used in \S\ref{sec:experiments}. Each app runs on a freshly reset emulator image with a dedicated benchmark account; login state and account-private data are persisted across tasks within the same app but never shared across apps.

\section{Prompts}
\label{app:prompts}

All prompts are reproduced verbatim. We use them with the model versions reported in Appendix~\ref{app:llm-usage}.

\subsection{Metadata extraction prompt (page summary, direct intent, route distillation)}
\label{app:prompt-summary}

The page summary, direct-intent extraction, and route distillation are performed by a single VLM call that extracts all three fields jointly. Given a sequence of annotated screenshots (with green rectangles marking tap targets and arrows marking scrolls) plus optional action descriptions, the VLM returns a JSON object with \texttt{visual\_page\_summary}, \texttt{intent\_from\_current\_screen}, and \texttt{navigation\_memory} (containing \texttt{relevant\_waypoint\_sequence} and \texttt{transition\_hints}).

\paragraph{System prompt\\}
\begin{promptbox}
You extract compact retrieval metadata and navigation memory for the final Android UI screen in a visual trajectory.

Return only valid JSON. No markdown. No explanation.

You are given:
1. a sequence of annotated screenshots
2. optional action descriptions that help explain the visual annotations

The screenshots are ordered from earlier to later. The last screenshot is the current/final screen. Only the last screenshot should be treated as the target screen for retrieval.

The earlier screenshots show how the user arrived at the final screen. They may contain visualized actions:
- scroll or swipe actions are shown with arrows
- other actions are shown with green rectangles around the clicked target

Most important rule for retrieval metadata: focus on the active interaction layer of the final screenshot.

Active interaction layer:
- If a modal, dialog, popup, bottom sheet, drawer, menu, permission prompt, overlay, snackbar action panel, or foreground panel is open, treat that foreground layer as the current screen.
- In that case, ignore the dimmed, blurred, partially covered, or background content behind it.
- Do not extract intents from the background screen when a foreground layer is active.
- If a bottom sheet or modal covers only part of the screen, still treat the sheet/modal as the active screen and ignore unrelated background content.
- If an overflow menu or floating menu is open, describe the menu's purpose and visible menu options, not the page behind it.
- If a permission dialog or system dialog is open, describe the dialog's intent, not the app screen behind it.
- If there is no foreground overlay, use the full final screenshot.

Important for retrieval metadata:
- The action annotations on earlier screenshots are context for the path.
- If the final screenshot contains visible actions, controls, tabs, menu items, or destinations that would lead somewhere else, do not treat those as the user's intent.
- Do not list sibling pages or destinations visible on the final screen as intents.
- Do not infer that the user wants to go to another page just because that option is visible.
- The goal is to describe what the active final screen/layer itself directly satisfies if it is the intended destination.

Extract these fields:

visual_page_summary:
- one short phrase describing the stable purpose of the active final screen or foreground layer
- base this mainly on the active layer in the final screenshot
- use earlier screenshots only to disambiguate what the final active layer is
- do not describe the path or previous screens
- if a modal/dialog/menu/sheet is active, summarize that foreground component, not the background page

intent_from_current_screen:
- user intents directly satisfied by the active final screen or foreground layer itself
- include information the user can inspect on the active layer
- include actions the user can complete on the active layer without needing to navigate to another screen
- do not include intents requiring leaving the active layer
- do not include sibling tabs, bottom navigation destinations, menu entries from the background, or links to other pages
- do not include low-level UI mechanics
- if an overlay is open, include only intents supported by the overlay, dialog, menu, sheet, or foreground panel

navigation_memory:
- a compact reusable memory describing how to reach the CURRENT/final node from the trajectory
- use earlier screenshots and action descriptions to infer the clean semantic route to the final screen
- the CURRENT node remains the source of truth
- remove loops, retries, oscillations, backtracking, and irrelevant detours
- prefer the shortest semantically valid forward route to the CURRENT node
- if early path segments vary or are noisy, prefer the shared sub-route closest to the CURRENT node
- keep everything strictly in-app
- never imply leaving the app, going to the launcher, or using the OS home screen
- if a waypoint could be confused with OS home, rewrite it as an in-app destination such as "home feed tab (in-app)", "main feed", or "app home screen"
- waypoints must be semantic page/state labels, not action phrases
- transition hints must describe semantic guidance, not low-level gesture mechanics
- avoid coordinates, bounding boxes, brittle visual details, and literal event-log replay
- prefer reusable semantic labels over usernames, product names, exact post topics, timestamps, counts, or one-off visible values unless essential
- do not mention internal reasoning, clusters, node ids, graph-analysis language, retrieval, ranking, screenshots, annotations, or trajectories

Output style rules:
- use lowercase for visual_page_summary and intent_from_current_screen
- navigation_memory may use title-case semantic waypoint labels if clearer
- keep only semantically distinct intents
- prefer specific user goals over generic descriptions
- ignore os/status bar data, ads, timestamps, temporary feed items, and promotional content
- avoid low-level UI words in intents: tap, click, press, icon, button, rectangle, arrow, swipe, scroll, tab, menu
- return at most 8 intents
- return at most 6 waypoints
- return at most 6 transition hints
\end{promptbox}

\paragraph{User prompt\\}
\begin{promptbox}
Extract retrieval metadata and navigation memory for the final/current screen.

Action descriptions, in trajectory order:
{action_descriptions}

Remember:
- screenshots are provided in order
- the last screenshot is the current/final screen
- earlier screenshots are only path context
- green rectangles and arrows show actions taken in the trajectory
- focus retrieval metadata only on the active interaction layer in the final screenshot
- if a modal, dialog, popup, bottom sheet, drawer, menu, permission prompt, or foreground overlay is open, extract metadata only from that foreground layer
- ignore dimmed, blurred, covered, or background content behind an active foreground layer
- do not treat actions available on the final screen as the user's intent unless the active final screen/layer itself directly satisfies that intent
- do not include destinations reachable from the final screen or from the background screen
- for navigation_memory, infer the cleanest semantic in-app route that reaches the current/final screen
- remove loops, retries, oscillations, backtracking, and irrelevant detours
- waypoints must be semantic page/state labels, not action phrases
- transition hints must be semantic guidance, not low-level gesture instructions

Return strict JSON only with exactly this schema:
{"visual_page_summary": "", "intent_from_current_screen": [],
 "navigation_memory": {"relevant_waypoint_sequence": [],
 "transition_hints": []}}
\end{promptbox}

\subsection{Target selector prompt}
\label{app:prompt-selector}

Given the user query and a list of candidate nodes, this prompt asks the VLM to rank the candidate nodes and select the single best target. Each candidate node has a \texttt{visual\_page\_summary} and \texttt{intent\_from\_current\_screen}.

\paragraph{System prompt\\}
\begin{promptbox}
You are selecting the best target node in an Android UI graph.

You receive:
1. a user query
2. a list of candidate UI nodes
3. each candidate has: index, visual_page_summary, intent_from_current_screen, optional retrieval scores

Your job: Choose the candidate whose CURRENT screen best satisfies the user query as the final target node.

Important ranking rules:
- Prefer the actual target screen, not a loading screen before it.
- Penalize loading screens, splash screens, blank screens, and temporary transition screens unless the user explicitly asks for them.
- Penalize modals, dialogs, bottom sheets, overlays, drawers, and overflow menus unless the user explicitly asks for that foreground component.
- Do not choose a page only because it has an option that leads to the target.
- Do not choose a settings page merely because it mentions the requested feature. Example: "timer sound settings" is not the same as "timer section".
- For page/navigation queries such as "navigate to the help page", "open timer", or "go to settings": prioritize visual_page_summary.
- For task/action queries such as "turn off alarm", "change timer sound", or "send feedback": use intent_from_current_screen more strongly.
- If the query asks for a general help page, prefer a help resources/support page over a specific help article, unless the query asks for that article.
- If the query asks for a specific help topic, prefer the matching help article.
- If a candidate is a loading screen for a page and another candidate is the actual loaded page/content, choose the actual loaded page/content.
- Choose only from the provided candidate indices.

Candidate labels:
- target: the current screen is the best final target.
- acceptable: relevant and usable, but not the best.
- transient_related: loading/temporary state related to the target.
- waypoint: can lead to the target but is not itself the target.
- related_but_wrong: related terms appear, but it is not the target.
- wrong: irrelevant.

Return only valid JSON. No markdown. No explanation outside JSON.

Output schema:
{"best_index": 0, "best_label": "target", "confidence": 0.0,
 "reason": "short reason", "ranked_candidates": [{"index": 0,
 "label": "target|acceptable|transient_related|waypoint|related_but_wrong|wrong",
 "score": 0.0, "reason": "short reason"}]}
\end{promptbox}

\paragraph{User prompt\\}
\begin{promptbox}
User query: {user_query}

Application description: {app_description}

Candidate UI nodes: {compact_candidates}

Select the best target node index.

Remember:
- The target is the node whose current screen itself satisfies the query.
- Do not choose loading screens if an actual content page is available.
- Do not choose menus, modals, or bottom sheets unless the query asks for them.
- Do not choose a settings page just because it mentions the requested feature.
- For "help page", prefer the general help/support/resources page over a specific article unless the query names that article.
\end{promptbox}

\subsection{Destination-memory injection template}
\label{app:prompt-injection}

At the beginning of task navigation, this text block is prepended to the backbone actor's system prompt with the memory plan to complete the current task. The injected memory contains the user instruction, the retrieved waypoint sequence, and the per-transition hints. The block is fixed across the trajectory in the open-loop canonical configuration; dynamic variants (Appendix~\ref{app:dynamic-variants}) update it during execution.

\begin{promptbox}
[Navigation Memory]
Final goal: {prompt}

Relevant waypoint sequence:
{waypoint_1} -> {waypoint_2} -> ... -> {waypoint_n}

Transition hints:
- {hint_1}
- {hint_2}
- ...
\end{promptbox}

\subsection{GPT-4o judge prompt}
\label{app:prompt-judge}

A GPT-4o call is used to compare the agent's final screenshot against the ground-truth target screenshot and user instruction to return a 1--5 ordinal score plus a one-sentence justification. Scores 4--5 are considered a success.

\begin{promptbox}
You are evaluating whether a mobile-app navigation agent reached the intended target page or target control for a user task.

Task: {prompt}

Image 1 is the AGENT's final screen.
Image 2 is the GROUND-TRUTH target screen.

Score from 1 to 5:
5 = exact target page/control reached.
4 = correct functional page/section reached; only minor visual, scroll, modal, or dynamic-content differences.
3 = partially correct: same broad area/app section, but wrong specific leaf, tab, field, or subpage.
2 = wrong section of the same app, or agent is clearly not at the requested destination.
1 = wrong app, launcher/home, stuck, or unrelated screen.

Be strict for score 4-5. A sibling settings page, nearby list item, or same broad menu but wrong specific target is score 3, not success.

Respond with ONLY a JSON object:
{"score": 1-5, "justification": "<brief explanation>"}
\end{promptbox}

\section{Additional Quantitative Results}
\label{app:additional-results}

\subsection{Per-app, per-backbone, per-benchmark breakdown}
\label{app:per-app}
Table~\ref{tab:per-app-destbench} (DestBench) and Table~\ref{tab:per-app-droidtask} (DroidTask-Nav and SPA-Bench-Nav) give the per-app results behind Table~\ref{tab:cross-backbone}; cells report successes over tasks for both the no-memory and +\method\ arms.

\begin{table}[!htbp]
\centering
\scriptsize
\setlength{\tabcolsep}{2pt}
\renewcommand{\arraystretch}{0.95}
\begin{tabularx}{\columnwidth}{@{}>{\raggedright\arraybackslash}X>{\centering\arraybackslash}p{0.13\columnwidth}>{\centering\arraybackslash}p{0.17\columnwidth}>{\centering\arraybackslash}p{0.13\columnwidth}>{\centering\arraybackslash}p{0.17\columnwidth}@{}}
\toprule
 & \multicolumn{2}{c}{\textbf{No memory}} & \multicolumn{2}{c}{\textbf{+ \method}} \\
\cmidrule(lr){2-3}\cmidrule(lr){4-5}
\textbf{App} & UI-TARS-1.5-7B & Qwen-VL-Max & UI-TARS-1.5-7B & Qwen-VL-Max \\
\midrule
Amazon      & 9/17 & 8/17 & 10/17 & 11/17 \\
Calendar    & 5/10 & 7/10 & 4/10  & 7/10  \\
Clock       & 1/9  & 4/9  & 3/9   & 7/9   \\
Google Maps & 12/24 & 12/24 & 15/24 & 17/24 \\
Settings    & 8/9  & 7/9  & 7/9   & 7/9   \\
YouTube     & 9/15 & 12/15 & 10/15 & 11/15 \\
\midrule
\textbf{Total} & 44/84 & 50/84 & 49/84 & 60/84 \\
\textbf{SR (\%)} & 52.4 & 59.5 & 58.3 & 71.4 \\
\bottomrule
\end{tabularx}
\caption{\textbf{DestBench per-app drilldown} (6 apps, 84 tasks). Human-annotated strict SR.}
\label{tab:per-app-destbench}
\end{table}

\begin{table*}[t]
\centering
\small
\setlength{\tabcolsep}{5pt}
\begin{tabular}{llrrrrl}
\toprule
App & Graph condition & Success & No memory & $\Delta$ pp & 95\% CI pp & Exact $p$ \\
\midrule
Amazon & Random & 11/17 & 9/17 & +11.8 & [+0.0, +29.4] & 0.500 \\
Calendar & Random & 7/10 & 5/10 & +20.0 & [-30.0, +60.0] & 0.688 \\
Clock & Random & 4/9 & 1/9 & +33.3 & [+0.0, +66.7] & 0.250 \\
Google Maps & Random & 16/24 & 12/24 & +16.7 & [-8.3, +41.7] & 0.344 \\
Settings & Random & 7/9 & 8/9 & -11.1 & [-33.3, +0.0] & 1.000 \\
YouTube & Random & 9/15 & 9/15 & +0.0 & [-33.3, +33.3] & 1.000 \\
Amazon & Auto-BFS & 10/17 & 9/17 & +5.9 & [+0.0, +17.6] & 1.000 \\
Calendar & Auto-BFS & 4/10 & 5/10 & -10.0 & [-30.0, +0.0] & 1.000 \\
Clock & Auto-BFS & 3/9 & 1/9 & +22.2 & [+0.0, +55.6] & 0.500 \\
Google Maps & Auto-BFS & 15/24 & 12/24 & +12.5 & [-16.7, +41.7] & 0.581 \\
Settings & Auto-BFS & 7/9 & 8/9 & -11.1 & [-33.3, +0.0] & 1.000 \\
YouTube & Auto-BFS & 10/15 & 9/15 & +6.7 & [-13.3, +26.7] & 1.000 \\
Amazon & Manual & 12/17 & 9/17 & +17.6 & [+0.0, +35.3] & 0.250 \\
Calendar & Manual & 6/10 & 5/10 & +10.0 & [-20.0, +40.0] & 1.000 \\
Clock & Manual & 3/9 & 1/9 & +22.2 & [+0.0, +55.6] & 0.500 \\
Google Maps & Manual & 20/24 & 12/24 & +33.3 & [+12.5, +54.2] & 0.021 \\
Settings & Manual & 8/9 & 8/9 & +0.0 & [+0.0, +0.0] & 1.000 \\
YouTube & Manual & 10/15 & 9/15 & +6.7 & [+0.0, +20.0] & 1.000 \\
\bottomrule
\end{tabular}
\caption{UI-TARS-1.5-7B per-app comparisons against no memory on DestBench (human strict labels; intervals and tests as in Appendix~\ref{app:historical-uncertainty}).}
\label{tab:historical-per-app-uncertainty}
\end{table*}
\begin{table*}[t]
\centering
\small
\setlength{\tabcolsep}{5pt}
\begin{tabular}{llrrrrl}
\toprule
App & Graph condition & Success & No memory & $\Delta$ pp & 95\% CI pp & Exact $p$ \\
\midrule
Amazon & Random & 9/17 & 8/17 & +5.9 & [-17.6, +29.4] & 1.000 \\
Calendar & Random & 8/10 & 7/10 & +10.0 & [-20.0, +40.0] & 1.000 \\
Clock & Random & 7/9 & 4/9 & +33.3 & [+0.0, +66.7] & 0.250 \\
Google Maps & Random & 12/24 & 12/24 & +0.0 & [-25.0, +25.0] & 1.000 \\
Settings & Random & 6/9 & 7/9 & -11.1 & [-44.4, +22.2] & 1.000 \\
YouTube & Random & 12/15 & 12/15 & +0.0 & [-26.7, +26.7] & 1.000 \\
Amazon & Auto-BFS & 11/17 & 8/17 & +17.6 & [+0.0, +35.3] & 0.250 \\
Calendar & Auto-BFS & 7/10 & 7/10 & +0.0 & [+0.0, +0.0] & 1.000 \\
Clock & Auto-BFS & 7/9 & 4/9 & +33.3 & [+0.0, +66.7] & 0.250 \\
Google Maps & Auto-BFS & 17/24 & 12/24 & +20.8 & [-8.3, +50.0] & 0.267 \\
Settings & Auto-BFS & 7/9 & 7/9 & +0.0 & [-33.3, +33.3] & 1.000 \\
YouTube & Auto-BFS & 11/15 & 12/15 & -6.7 & [-33.3, +20.0] & 1.000 \\
Amazon & Manual & 8/17 & 8/17 & +0.0 & [-29.4, +29.4] & 1.000 \\
Calendar & Manual & 7/10 & 7/10 & +0.0 & [-30.0, +30.0] & 1.000 \\
Clock & Manual & 6/9 & 4/9 & +22.2 & [+0.0, +55.6] & 0.500 \\
Google Maps & Manual & 16/24 & 12/24 & +16.7 & [-8.3, +41.7] & 0.344 \\
Settings & Manual & 7/9 & 7/9 & +0.0 & [-44.4, +44.4] & 1.000 \\
YouTube & Manual & 12/15 & 12/15 & +0.0 & [-26.7, +26.7] & 1.000 \\
\bottomrule
\end{tabular}
\caption{Qwen-VL-Max per-app comparisons against no memory on DestBench (human strict labels; intervals and tests as in Appendix~\ref{app:historical-uncertainty}).}
\label{tab:qwen-per-app-uncertainty}
\end{table*}

\begin{table*}[t]
\centering
\small
\setlength{\tabcolsep}{5pt}
\begin{tabular}{llrrrrl}
\toprule
Benchmark & Backbone & Success & No memory & $\Delta$ pp & 95\% CI pp & Exact $p$ \\
\midrule
DroidTask-Nav & UI-TARS-1.5-7B & 107/150 & 95/150 & +8.0 & [+0.0, +16.0] & 0.073 \\
DroidTask-Nav & Qwen-VL-Max & 106/150 & 99/150 & +4.7 & [-2.0, +12.0] & 0.265 \\
SPA-Bench-Nav & UI-TARS-1.5-7B & 23/30 & 20/30 & +10.0 & [-10.0, +30.0] & 0.508 \\
SPA-Bench-Nav & Qwen-VL-Max & 28/30 & 23/30 & +16.7 & [+0.0, +33.3] & 0.125 \\
\bottomrule
\end{tabular}
\caption{Converted-benchmark comparisons against no memory (human strict labels; intervals and tests as in Appendix~\ref{app:historical-uncertainty}).}
\label{tab:converted-paired}
\end{table*}

\newpage
\subsection{Per-app uncertainty}
\label{app:historical-uncertainty}
Tables~\ref{tab:historical-per-app-uncertainty} and~\ref{tab:qwen-per-app-uncertainty} compare each graph condition with no memory per app on the 84 DestBench tasks, and Table~\ref{tab:converted-paired} does the same for the converted benchmarks. All outcomes are human strict labels. Intervals are 95\% paired bootstrap intervals over tasks (10,000 resamples), and $p$ values come from exact two-sided McNemar tests; neither is adjusted for multiple comparisons, here or in Tables~\ref{tab:cross-backbone} and~\ref{tab:original-paired-results}.

\subsection{Graph-arm memory provenance}
\label{app:graph-arm-provenance}
In some random- and manual-arm app cells, the harness supplied the same app's memory from a different graph condition. Table~\ref{tab:graph-arm-clean} repeats the comparisons of Table~\ref{tab:original-paired-results} without those apps; every gain remains positive.
\begin{table}[!htbp]
\centering
\scriptsize
\setlength{\tabcolsep}{2.5pt}
\resizebox{\columnwidth}{!}{%
\begin{tabular}{@{}lrrrrr@{}}
\toprule
Arm (excluded apps: memory received) & $n$ & No mem. & Arm & $\Delta$ pp [95\% CI] & $p$ \\
\midrule
UI-TARS-1.5-7B random (YouTube, Settings: Auto-BFS) & 60 & 27 & 38 & +18.3 [5.0, 31.7] & 0.027 \\
UI-TARS-1.5-7B manual (Clock: random-graph targets) & 75 & 43 & 56 & +17.3 [8.0, 28.0] & 0.002 \\
Qwen-VL-Max random (Settings: Auto-BFS) & 75 & 43 & 48 & +6.7 [$-6.7$, 20.0] & 0.424 \\
Qwen-VL-Max manual (Clock: random-graph targets) & 75 & 46 & 50 & +5.3 [$-8.0$, 18.7] & 0.557 \\
\bottomrule
\end{tabular}}
\caption{\textbf{Graph conditions without the affected apps.} Each row names the excluded apps and the memory they received. Human strict success; intervals and tests as in Appendix~\ref{app:historical-uncertainty}.}
\label{tab:graph-arm-clean}
\end{table}

\subsection{Per-app graph-quality breakdown}
\label{app:per-app-graph-quality}
Table~\ref{tab:per-app} is the per-app counterpart to Table~\ref{tab:graph-quality}. Per-app patterns make the aggregate finding concrete: deep-menu apps benefit most from any memory (the destination is several hops below the entry screen), while shallow apps like Settings and Calendar benefit least because their residual difficulty is execution rather than navigation.

\begin{table*}[tp]
\centering
\scriptsize
\setlength{\tabcolsep}{4pt}
\renewcommand{\arraystretch}{0.95}
\begin{tabularx}{0.85\linewidth}{@{}>{\raggedright\arraybackslash}X>{\centering\arraybackslash}p{0.12\linewidth}>{\centering\arraybackslash}p{0.12\linewidth}>{\centering\arraybackslash}p{0.12\linewidth}>{\centering\arraybackslash}p{0.12\linewidth}@{}}
\toprule
\textbf{App} & \textbf{No memory} & \textbf{Random} & \textbf{Auto-BFS} & \textbf{Manual} \\
\midrule
Amazon & 9/17 & 11/17 & 10/17 & 12/17 \\
Calendar & 5/10 & 7/10 & 4/10  & 6/10 \\
Clock & 1/9 & 4/9 & 3/9 & 3/9 \\
Google Maps & 12/24 & 16/24 & 15/24 & 20/24 \\
Settings & 8/9 & 7/9 & 7/9 & 8/9 \\
YouTube & 9/15 & 9/15 & 10/15 & 10/15 \\
\midrule
Total & 44/84 & 54/84 & 49/84 & 59/84 \\
Success Rate & 52.4\% & 64.3\% & 58.3\% & 70.2\% \\
\bottomrule
\end{tabularx}
\caption{\textbf{Per-app graph-quality breakdown on \benchmark using UI-TARS-1.5-7B.} Cells are \texttt{successes / attempts} under the strict human-scoring rule; all cells are human-labeled (Appendix~\ref{app:historical-uncertainty}).}
\label{tab:per-app}
\end{table*}

\subsection{Route cleanliness}
\label{app:route-cleanliness}
Table~\ref{tab:route-cleanliness-correlations} reports how properties of the injected route correlate with success in the 313 GPT-4o-judged runs of Section~\ref{sec:exp-graph-quality}. Routes from manual graphs are shorter than Auto-BFS routes (2.9 vs.\ 6.0 waypoints on average).

\begin{table}[t]
\centering\setlength{\tabcolsep}{4pt}
\small
\begin{tabular}{lrrr}
\toprule
Injected-route measure & $n$ & Point-biserial $r$ & $p$ \\
\midrule
Route length (waypoints) & 313 & $-0.16$ & 0.006 \\
Transition-hint count & 313 & $-0.19$ & 0.001 \\
Loop count & 313 & $-0.03$ & 0.652 \\
Backtrack count & 313 & $+0.04$ & 0.467 \\
Detour count & 313 & $+0.02$ & 0.783 \\
\bottomrule
\end{tabular}
\caption{\textbf{Route-cleanliness associations.} GPT-4o-judged runs of Section~\ref{sec:exp-graph-quality}. Route length and hint count correlate negatively with success; this observational analysis does not establish causality.}
\label{tab:route-cleanliness-correlations}
\end{table}

\subsection{Memory format and retrieval-depth ablation}
\label{app:memory-format}
Ablations supporting the canonical configuration: waypoint vs step-numbered memory crossed with top-$k$ retrieval depth. Step-numbered observations trigger spurious \texttt{wait()} loops when the agent treats a state observation as an event; top-$k>1$ multi-route injection pollutes context with routes that do not match the live UI. The canonical configuration is semantic waypoints with top-1 retrieval.

\subsection{Cross-family and prior-memory outcomes}
\label{app:cross-family-outcomes}
The AgentCPM-GUI rows in Table~\ref{tab:main-result-cost} are paired runs on the 84 DestBench tasks, with later app versions than the UI-TARS-1.5-7B and Qwen-VL-Max rows and labels blind to memory condition; memory raises success from 28/84 to 33/84 (exact McNemar $p=0.23$). Some of these episodes began from in-app state left by the previous task, such as Clock's last tab or a Settings sub-page. The MAI-UI rows come from a separate 79-task six-app evaluation, 72 of whose tasks overlap with DestBench; three baseline tasks were not run and are counted as failures, and one baseline task lacks a label, so the two rates use different denominators and are not a paired comparison. The PG-Agent adaptation reaches 36/84 strict successes (37/84 under a second human labeling), and GUI-Explorer reaches 47/84 with a GPT-4o actor and its released knowledge base.

\subsection{Cost analysis}
\label{app:cost}
Table~\ref{tab:cost-analysis} reports the one-time exploration effort for each core app. A walk is one exploration episode from the app root (Appendix~\ref{app:local-bfs}), and a forward action is an action that moves the explorer to a new or known screen; resets and undo steps are not counted. The explorer records both counts in each graph.

\begin{table}[t]
\centering\small
\begin{tabular}{lrr}
\toprule
Core app & Walks & Forward actions \\
\midrule
Maps & 869 & 3,166 \\
Amazon & 804 & 2,440 \\
YouTube & 638 & 2,174 \\
Clock & 994 & 3,990 \\
Settings & 148 & 510 \\
Calendar & 209 & 655 \\
\midrule
Total & 3,662 & 12,935 \\
\bottomrule
\end{tabular}
\caption{\textbf{Exploration effort per core app.} Walks and forward actions recorded during graph construction (Appendix~\ref{app:cost}).}
\label{tab:cost-analysis}
\end{table}

\subsection{Record-content ablation}
\label{app:response-period}
On 27 Instagram and eBay tasks with graphs and apps at matched versions, we vary which fields of the destination record are injected. In this cohort, transition hints alone gave the highest success (Table~\ref{tab:record-content-27}); Appendix~\ref{app:version-robustness} evaluates the same tasks after app updates.
\begin{table}[t]
\centering
\small
\begin{tabular}{lrrr}
\toprule
Record condition & Instagram & eBay & Total \\
\midrule
Transition hints & 4/11 & 13/16 & 17/27 \\
Full record & 2/11 & 12/16 & 14/27 \\
No memory & 4/11 & 10/16 & 14/27 \\
Route only & 2/11 & 11/16 & 13/27 \\
Goal only & 2/11 & 10/16 & 12/27 \\
\bottomrule
\end{tabular}
\caption{\textbf{Version-matched record-content ablation.} Human-judged Instagram/eBay tasks at matched graph versions. The small sample is suggestive.}
\label{tab:record-content-27}
\end{table}

\subsection{Informed no-memory baseline}
\label{app:informed-baseline}
To test whether structured records add value beyond general app knowledge, we compare them with a fixed, hand-written manual for each core app that gives general guidance but no task-specific destination or route (Table~\ref{tab:app-manual-86}). The manual and every record variant improve on no memory by four to nine tasks, and none of the differences is significant: app-specific text captures part of the benefit of destination memory, and this diagnostic does not separate it from the structured records.
\begin{table}[!htbp]
\centering
\scriptsize
\begin{tabular}{@{}lrrr@{}}
\toprule
Condition & Success & $\Delta$ & $p$ \\
\midrule
No memory & 35/80 & -- & -- \\
Goal-only record & 39/80 & +4 & 0.388 \\
Hand-written app manual & 41/80 & +6 & 0.109 \\
Full record & 39/80 & +4 & 0.541 \\
Route-only record & 42/80 & +7 & 0.167 \\
Transition-hints record & 44/80 & +9 & 0.108 \\
\bottomrule
\end{tabular}
\caption{\textbf{Informed no-memory baseline.} UI-TARS-1.5-7B, human-judged strict success on 80 core-app tasks. The app manual gives fixed app-level guidance without task-specific destinations or routes. $p$: exact two-sided McNemar against no memory.}
\label{tab:app-manual-86}
\end{table}

\section{Format-Adapted Reimplementation Protocol}
\label{app:reimpl}

\subsection{Per-system adaptation rules}
\label{app:reimpl-rules}
Table~\ref{tab:reimpl-rules} lists how each prior system's memory abstraction was reconstructed on our hypernode graph for Table~\ref{tab:main-result-cost}, and what was deliberately not reproduced (custom actors, validators, planners).

\begin{table}[!htbp]
\centering
\scriptsize
\setlength{\tabcolsep}{3pt}
\renewcommand{\arraystretch}{1.05}
\begin{tabularx}{\columnwidth}{@{}p{0.18\columnwidth}p{0.16\columnwidth}X@{}}
\toprule
\textbf{System} & \textbf{Backbone} & \textbf{Protocol adaptation} \\
\midrule
GUI-Explorer   & GPT-4o & Used their published interaction-triple KB, retrieval, and guideline-generation pipeline on our emulator/app versions; exploration was not rerun. \\
PG-Agent       & Qwen2.5-VL-72B & Built a page graph from \benchmark\ task episodes with their construction pipeline, retrieved page guidelines by screen embedding, and replaced the multi-agent planner with a single Qwen2.5-VL-72B navigator. \\
\bottomrule
\end{tabularx}
\caption{\textbf{Evaluation protocol for prior systems on \benchmark.} Each row keeps the system's memory abstraction but uses our tasks and human-annotated strict-SR judge.}
\label{tab:reimpl-rules}
\end{table}

\section{Qualitative Examples}
\label{app:qualitative}

\subsection{Failure trajectories per class}
\label{app:failure-trajectories}
\begin{table}[!htbp]
\centering
\scriptsize
\setlength{\tabcolsep}{2pt}
\renewcommand{\arraystretch}{0.9}
\begin{tabular}{lrrl}
\toprule
\textbf{Class} & \textbf{N} & \textbf{\%} & \textbf{Addressed by} \\
\midrule
A: wrong retrieval    & 22 & 46 & embeddings, coverage \\
B: near miss          & 15 & 31 & target selection \\
C: execution failure  & 11 & 23 & step budget, replanning \\
\bottomrule
\end{tabular}
\caption{\textbf{Failure taxonomy.} Labeled open-loop \benchmark\ failures.}
\label{tab:failure-modes}
\end{table}

\begin{figure*}[t]
\centering
\includegraphics[height=5.5cm]{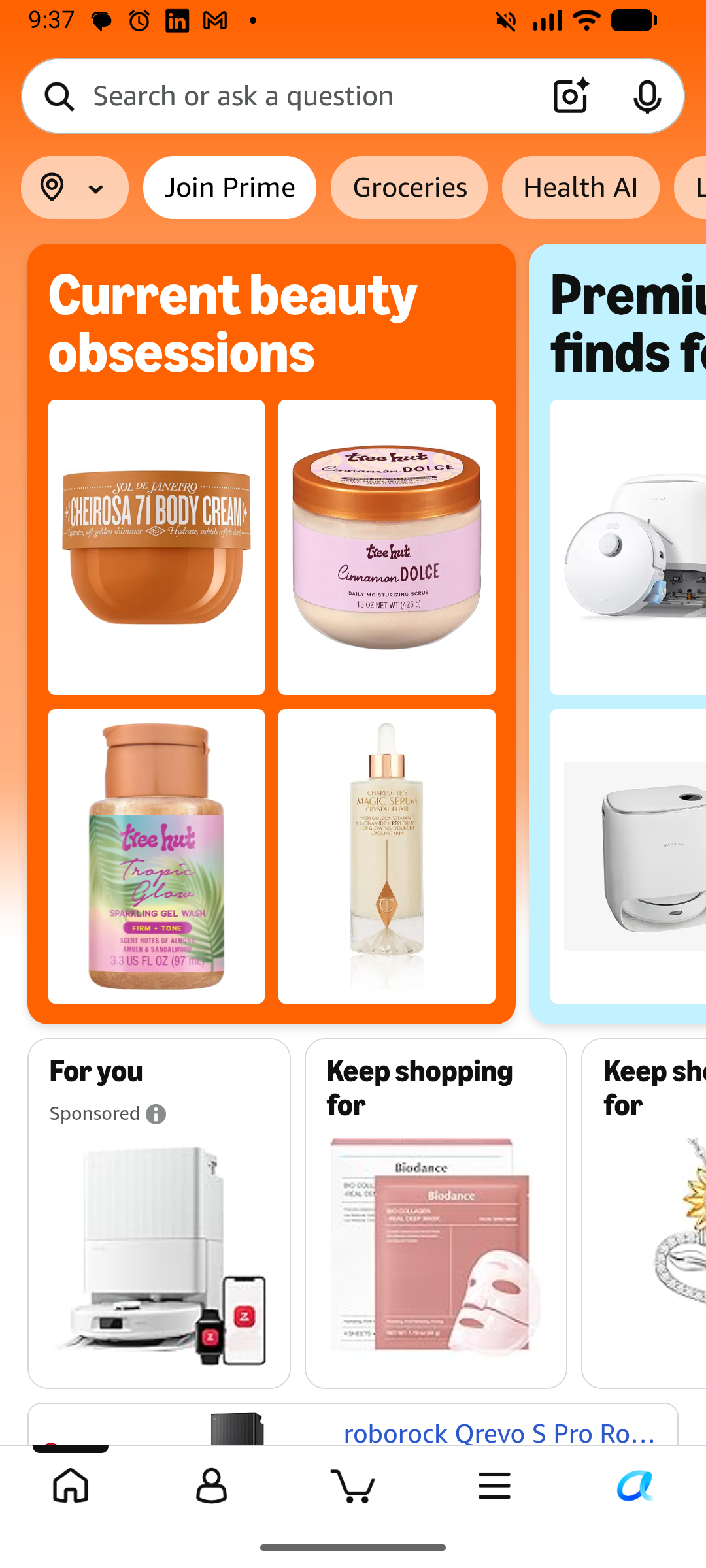}\hspace{4pt}
\includegraphics[height=5.5cm]{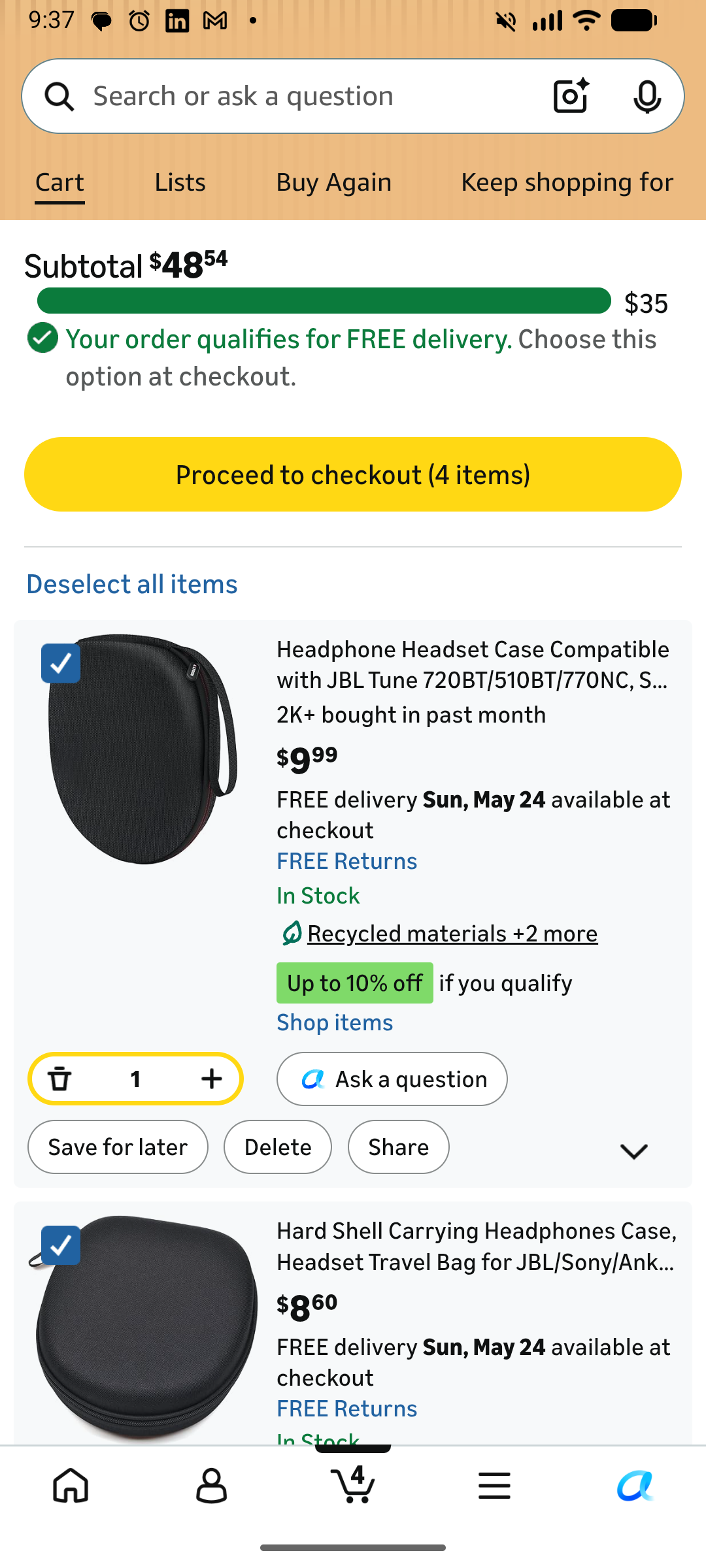}\hspace{4pt}
\includegraphics[height=5.5cm]{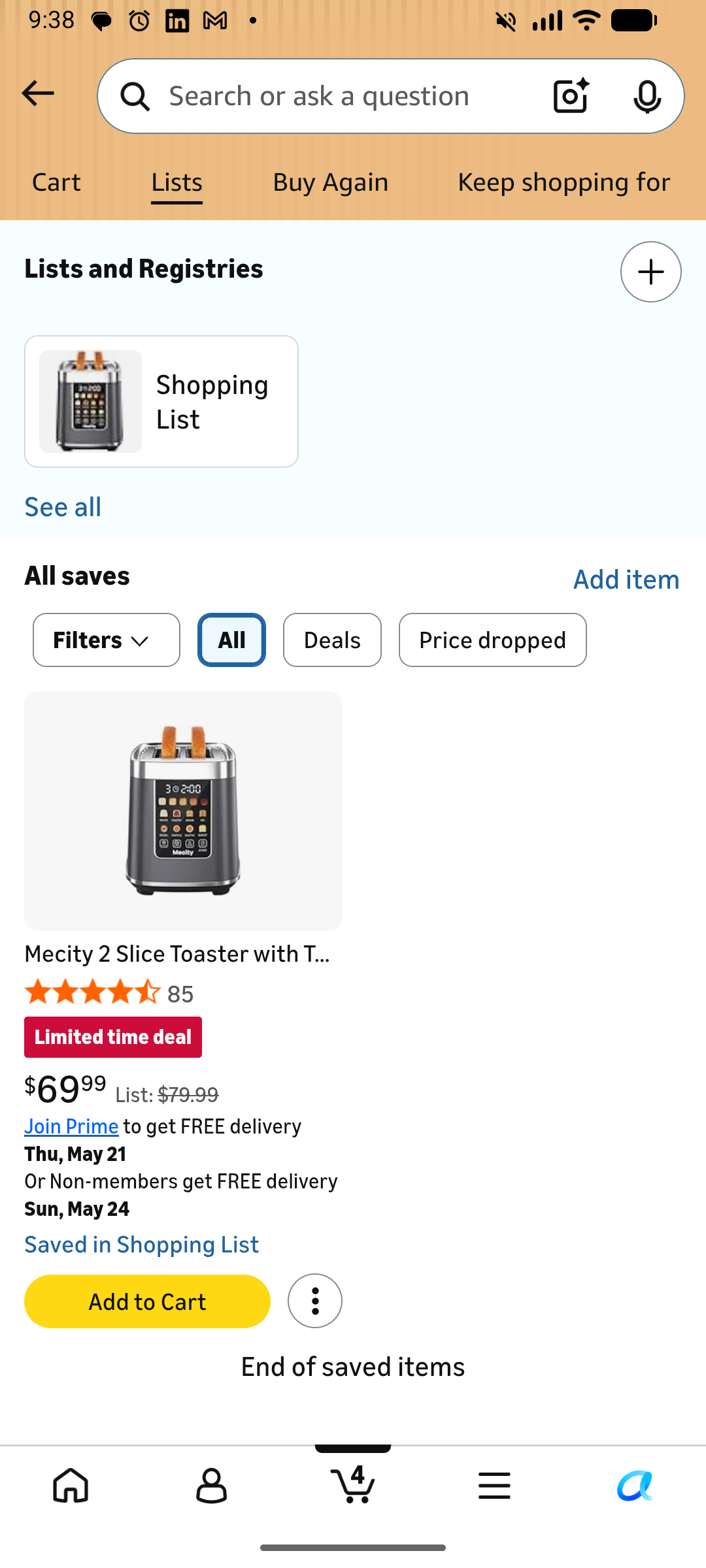}
\caption{\textbf{Class A Failure (Wrong retrieval)} Task: ``Visit the user's saved cards'' (Amazon).
Memory: \texttt{Shopping Cart $\rightarrow$ Lists Tab $\rightarrow$ Lists and Registries}.
The retrieval system incorrectly matched the task to the Lists node. The agent then followed the memory faithfully: (1)~Started at Amazon home page, (2)~Navigated to Shopping Cart page and tapped the Lists tab, (3)~Arrived at the Lists and Registries page showing a Shopping List. The embedding confused ``saved cards'' from the task with ``saved items/lists.''}
\label{fig:fail-a}
\end{figure*}

\begin{figure*}[t]
\centering
\includegraphics[height=5.5cm]{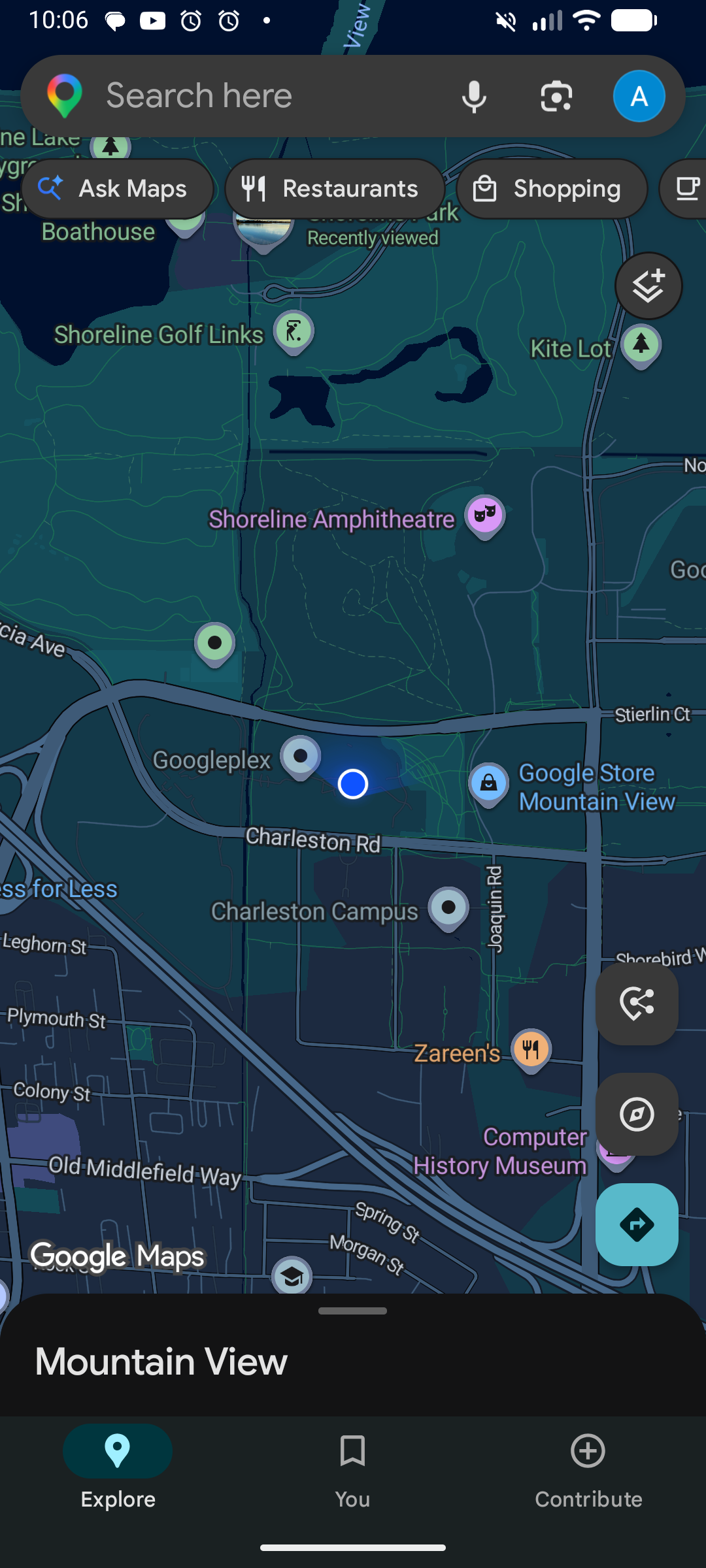}\hspace{4pt}
\includegraphics[height=5.5cm]{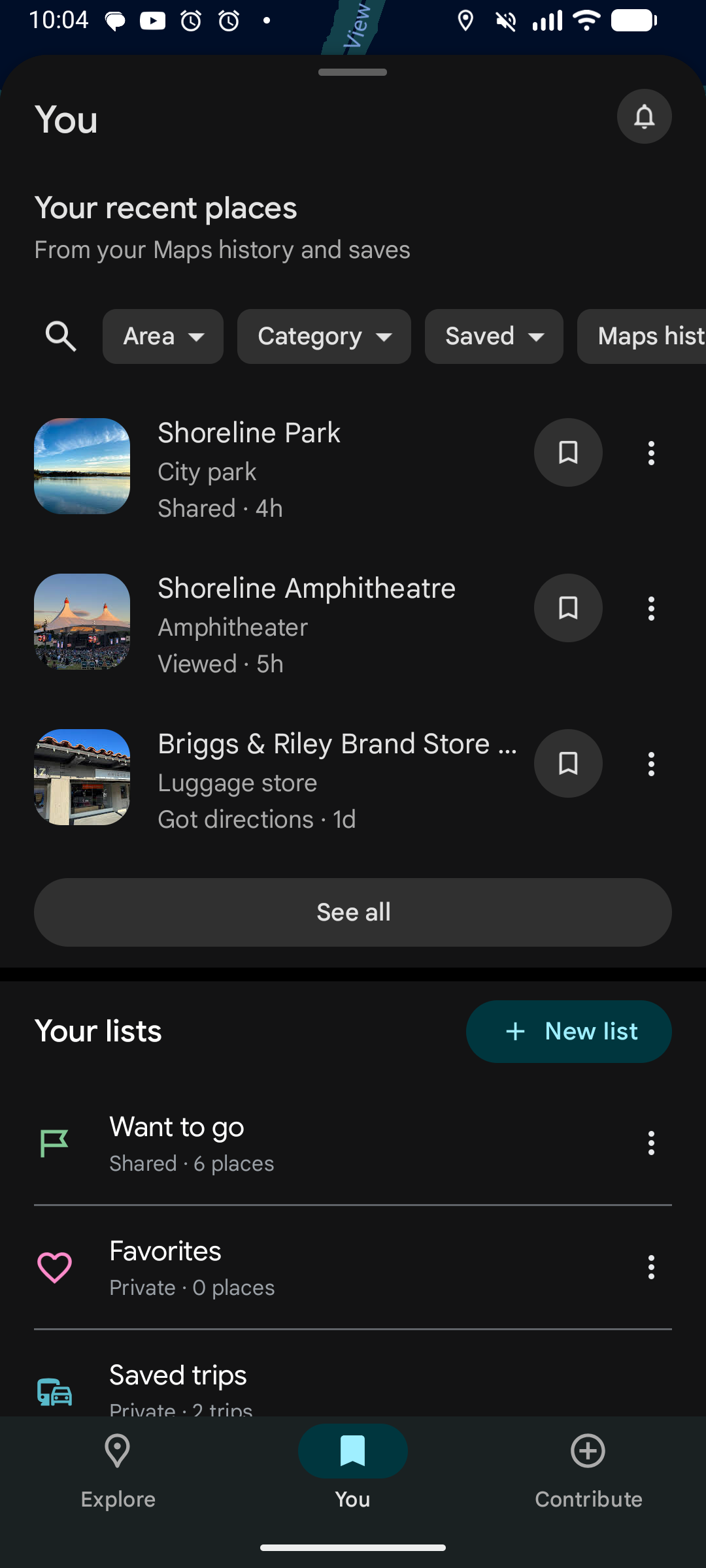}\hspace{4pt}
\includegraphics[height=5.5cm]{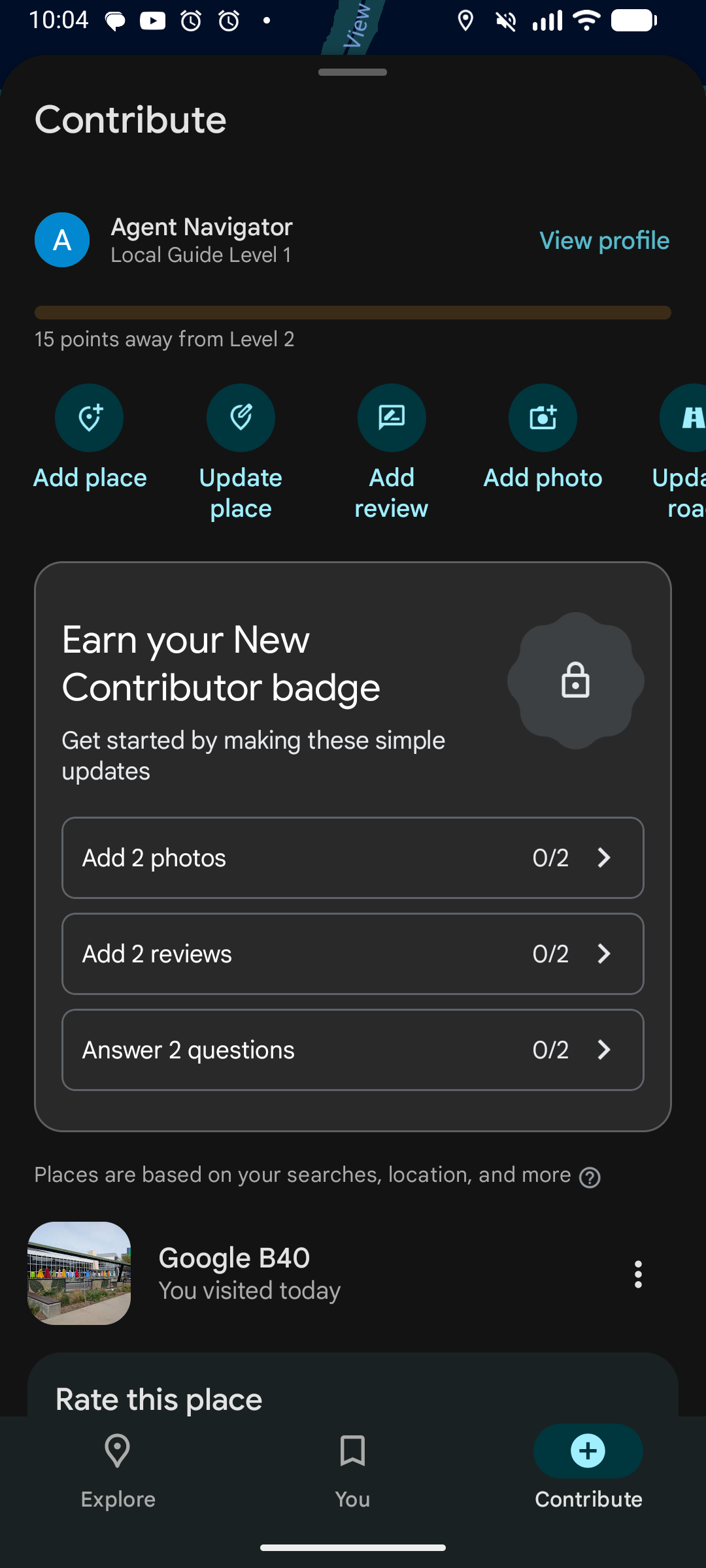}
\caption{\textbf{Class B Failure (Near miss)} Task: ``View the user's contribution profile'' (Google Maps). 
Memory: \texttt{Google Maps Home $\rightarrow$ Contribute hub $\rightarrow$ Profile details}.
The agent followed the memory: (1)~Started on the Maps home page, (2)~Tapped the Contribute tab, (3)~Arrived at the Contribute page showing the ``View profile'' link. It then declared success without clicking it. The retrieval selected the Contribute hub node, which is the parent node to the correct Profile details node.}
\label{fig:fail-b}
\end{figure*}

\begin{figure*}[t]
\centering
\includegraphics[height=5.5cm]{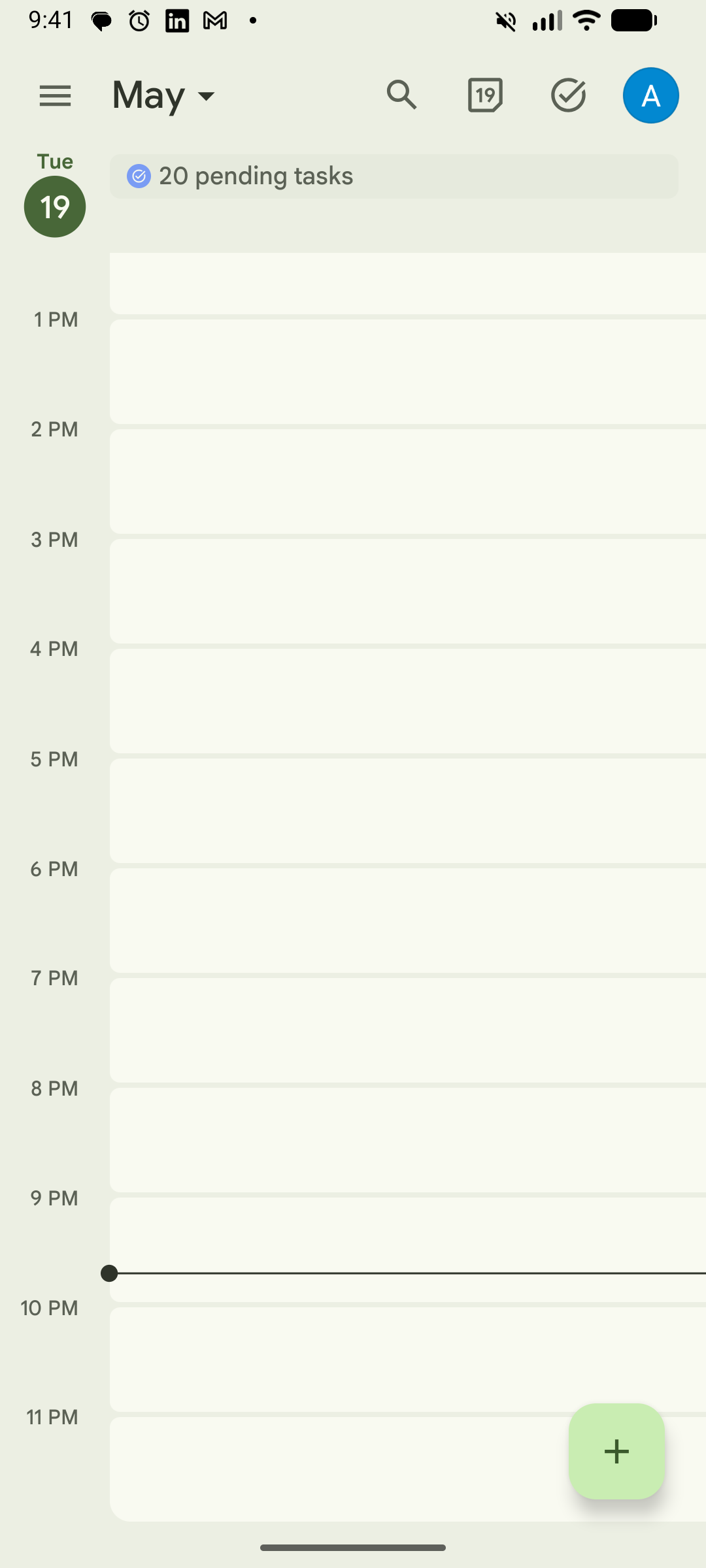}\hspace{4pt}
\includegraphics[height=5.5cm]{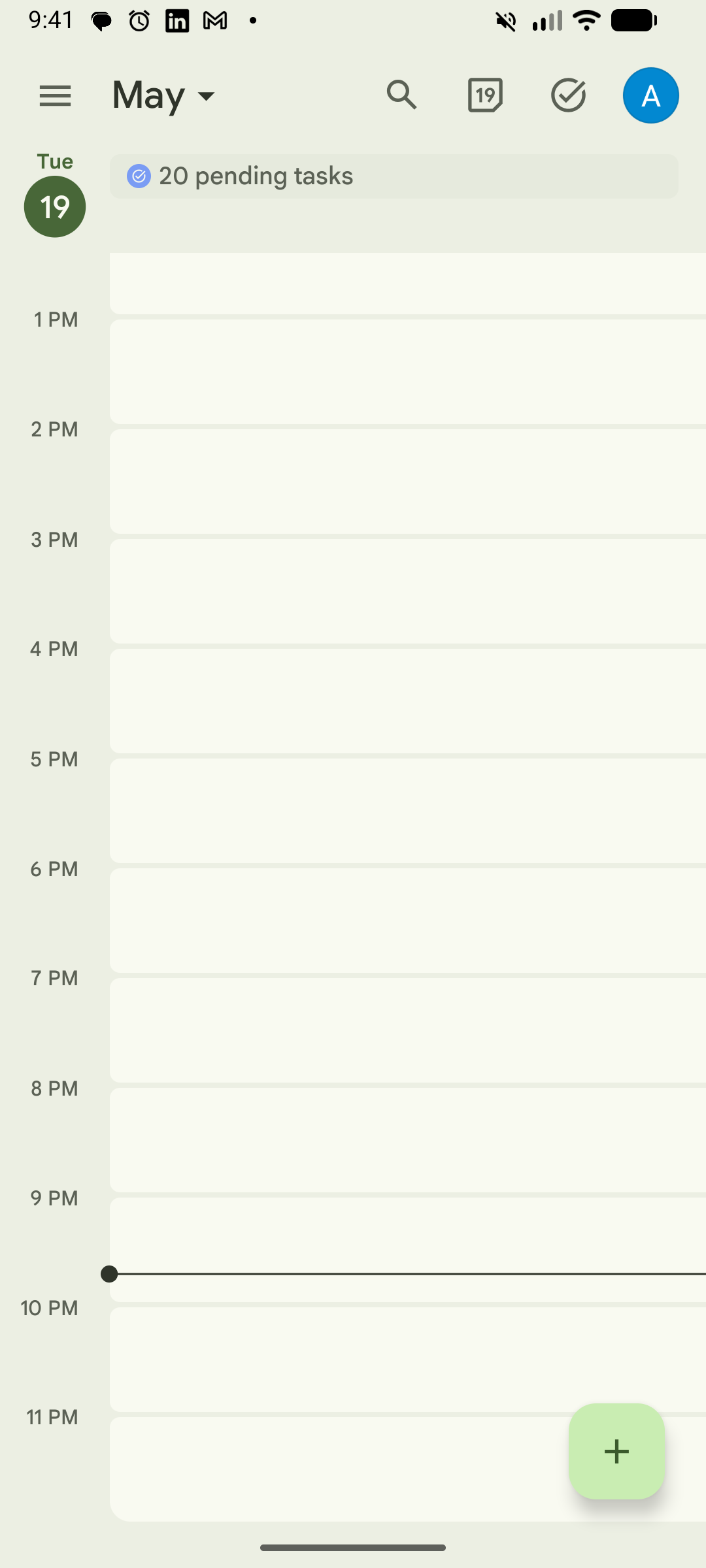}\hspace{4pt}
\includegraphics[height=5.5cm]{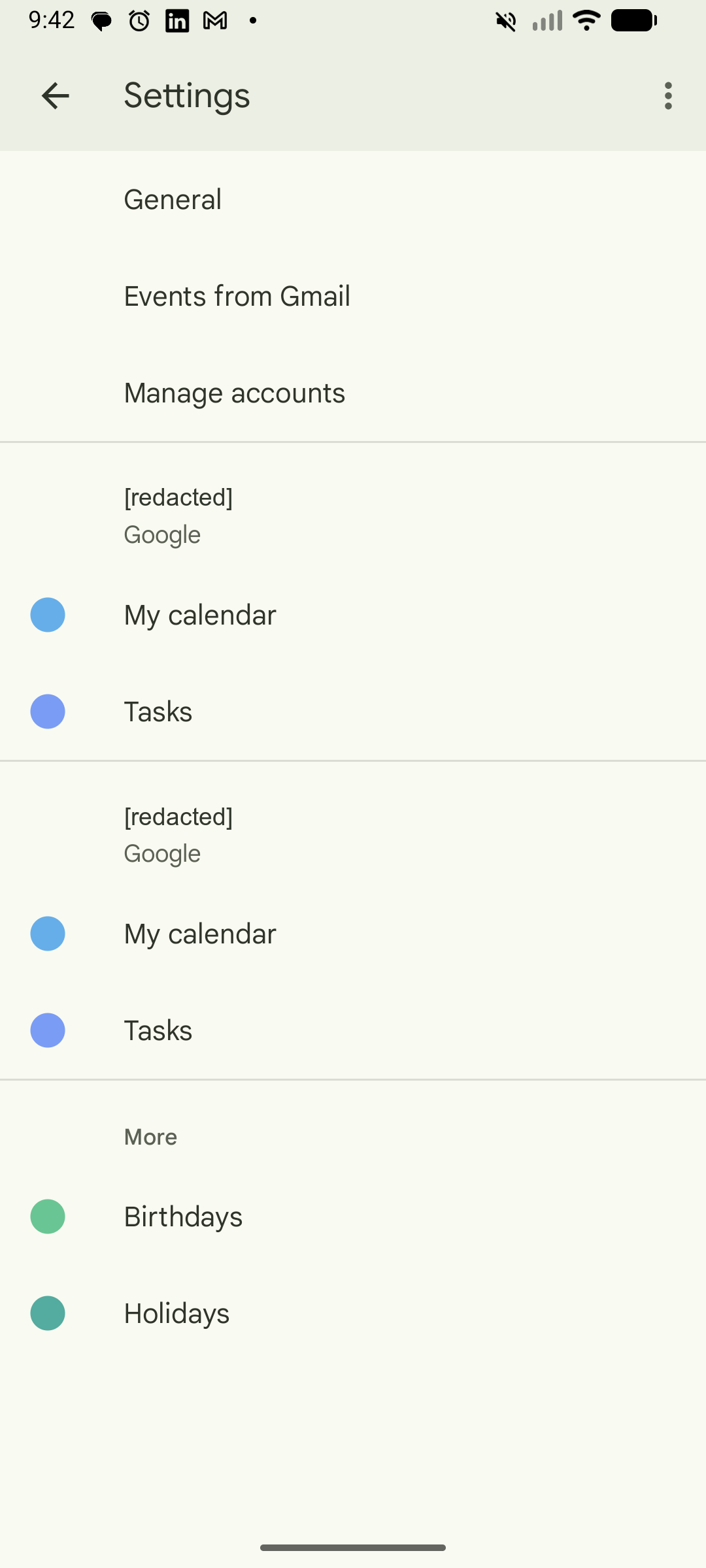}
\caption{\textbf{Class C Failure (Execution failure)}
Task: ``Toggle Use device time zone'' (Calendar).
Memory: \texttt{Calendar Day View $\rightarrow$ Side Navigation Menu $\rightarrow$ Settings Page} (correct).
The agent followed the memory: (1)~Started on the Calendar day view, (5)~Remained on the same Calendar view after 4 wasted steps, and (15)~Finally reached the Settings page but exhausted the step budget ($T{=}15$) before locating the specific toggle. The memory was correct but the agent could not efficiently navigate the multi-level settings hierarchy. Account identifiers in the final screenshot are redacted.}
\label{fig:fail-c}
\end{figure*}

Figures~\ref{fig:fail-a}--\ref{fig:fail-c} show one canonical trajectory per failure class from Table~\ref{tab:failure-modes}. \textbf{Class~A} (wrong retrieval, 46\%): the embedding-based retrieval matched the query to the wrong graph node, and the agent faithfully followed incorrect waypoints to a semantically-related but wrong destination. \textbf{Class~B} (near miss, 31\%): the agent reached the correct area of the app but stopped at a parent page, sibling leaf, or adjacent sub-item rather than the exact target. \textbf{Class~C} (execution failure, 23\%): the memory pointed in the right direction but the agent could not execute the navigation within the step budget, typically looping or scrolling without locating the specific UI element.

\subsection{Success trajectories}
\label{app:success-trajectories}
Two representative successes illustrate how the destination block guides the actor. On Calendar, an unseen target buried four taps under a settings drawer is reached by following the waypoint sequence verbatim; the actor never opens the side-menu it would have explored without memory. On Google Maps, a multi-route destination (reachable through search or through the saved-places tab) is solved by the shorter cleaned route stored in the record; the per-transition hints disambiguate a tab whose icon collides with a sibling control.

\subsection{Raw destination-record examples}
\label{app:record-examples}
Each record stores a page summary, a list of direct intents, a waypoint sequence from the app root, and per-transition hints. A representative record for the YouTube target ``open the watch-history list'' contains the summary ``screen listing previously watched videos with per-item remove buttons,'' the intent ``view watch history,'' the waypoint sequence \texttt{Home $\to$ Library $\to$ History}, and the hint ``Library is the bottom-bar icon labeled You, not the side-drawer entry of the same name.'' Records are stored as JSONL and never modified after exploration.

\section{Detailed Related-Work Positioning}
\label{app:related-positioning}

Table~\ref{tab:related-detailed} expands the categorical positioning in Table~\ref{tab:related}, using concise fields so the comparison stays readable in the appendix flow.

\begin{table}[!t]
\centering
\scriptsize
\setlength{\tabcolsep}{3pt}
\renewcommand{\arraystretch}{1.1}
\begin{tabularx}{\columnwidth}{@{}>{\raggedright\arraybackslash}p{0.20\columnwidth}>{\raggedright\arraybackslash}X>{\raggedright\arraybackslash}p{0.22\columnwidth}>{\raggedright\arraybackslash}p{0.18\columnwidth}@{}}
\toprule
\textbf{System} & \textbf{Stored memory} & \textbf{Retrieved unit} & \textbf{Coupling} \\
\midrule
AutoDroid / AppAgent & UI graph + element notes & Element hint / note & Own LLM \\
KG-RAG & UTG distilled into KG store & Intent--trajectory pair & Own agent loop \\
GUI-explorer & Interaction-triple store & Current-screen cue & Own MLLM \\
GUI-Xplore & Video transition graph & App-level prior & Own VLM \\
PG-Agent / GraphPilot & Page / app knowledge graph & Guideline or action rule & Planner + validator \\
\midrule
\textbf{\method} & \textbf{Nav.\ graph + dest.\ records} & \textbf{Target page + compact route} & \textbf{Unmodified actor} \\
\bottomrule
\end{tabularx}
\caption{\textbf{Detailed positioning of \method.} Prior systems retrieve hints, trajectories, guidelines, or action rules and are coupled to their own actors; \method\ retrieves a structured destination record for an unchanged stepwise actor.}
\label{tab:related-detailed}
\end{table}

\section{Limitations, Robustness, and Ethics}
\label{app:limits}

\subsection{Dynamic-screen handling}
\label{app:dynamic-screens}
\benchmark\ screens contain dynamic content: advertisements, recommendation feeds, trending lists, and time-varying notifications. The metadata-extraction prompt (Appendix~\ref{app:prompt-summary}) is written to focus on the persistent active layer, and the judge prompt (Appendix~\ref{app:prompt-judge}) is told to tolerate cosmetic differences. Persistent dynamic state (e.g., login-gated content, A/B layouts) is a known failure mode and is documented per-app in our records.

\subsection{Cross-app-version robustness}
\label{app:version-robustness}
Graphs are tied to a specific app version on a specific device configuration. In a limited 27-task probe on Instagram (about five months between versions) and eBay (about ten weeks), full-record memory matched no memory at both versions (Table~\ref{tab:drift-probe}). This probe does not establish general cross-version robustness. Apps with sufficiently large navigation-affecting changes still require the graph to be revalidated or rebuilt before its records can be trusted.
\begin{table}[t]
\centering
\small
\begin{tabular}{llrr}
\toprule
App & Condition & Matched & Updated \\
\midrule
Instagram & No memory & 4/11 & 1/11 \\
 & Full record & 2/11 & 2/11 \\
 & Transition hints & 4/11 & 4/11 \\
\midrule
eBay & No memory & 10/16 & 10/16 \\
 & Full record & 12/16 & 9/16 \\
 & Transition hints & 13/16 & 11/16 \\
\midrule
Total & No memory & 14/27 & 11/27 \\
 & Full record & 14/27 & 11/27 \\
 & Transition hints & 17/27 & 15/27 \\
\bottomrule
\end{tabular}
\caption{\textbf{Limited stale-graph probe.} Human-judged paired tasks using the original graph at matched and updated app versions. Instagram: 416.0.0.47.66 to 437.0.0.33.78; eBay: 6.242.0.2 to 6.251.0.1. Full-record memory matched no memory in aggregate at both versions, with differing per-app outcomes; this small study does not establish general cross-version robustness.}
\label{tab:drift-probe}
\end{table}

\subsection{Multi-path and ambiguous-target handling}
\label{app:multipath}
When a destination is reachable by multiple distinct routes (e.g., side-menu vs bottom-tab), the record stores all routes that the explorer found and the top-1 retrieval selects the shortest cleaned route. When an instruction admits multiple visually-distinct targets, the selector prompt (Appendix~\ref{app:prompt-selector}) breaks ties using the five preference rules; if all preferences are tied, the lower-depth target wins.

\subsection{Privacy and sensitive-screen handling}
\label{app:privacy}
We use dedicated benchmark accounts for each app, with no real user data. Screens that would contain personal data on a real account (e.g., contacts, message bodies, payment instruments) are blank or contain synthetic placeholders. Any public release of screenshots requires screening for identifying content.

\subsection{Use of AI in paper preparation}
\label{app:ai-paper-usage}
LLM-based assistants were used for literature and manuscript review, drafting, LaTeX editing, and analysis scripts. The authors are responsible for verifying all statements, evidence, and calculations and for approving the final manuscript. The assistants are not authors.

\subsection{Models used in the \method\ pipeline}
\label{app:llm-usage}
All hosted models were called through Alibaba Cloud's DashScope API, except the GPT-4o judge, which used the OpenAI API. Exploration runs UI-TARS-1.5-7B locally for action grounding, uses \texttt{qwen3.6-plus}~\citep{qwen2026qwen36} to extract interactive elements, and uses \texttt{qwen3.6-flash} or UI-TARS-1.5-7B for loading-screen checks. Hypernode matching uses CLIP ViT-B/32 screenshot embeddings and \texttt{text-embedding-v4} page-layout embeddings (Appendix~\ref{app:hypernode-match}). Offline post-processing (page summary, direct intents, waypoint route, and transition hints) makes one \texttt{qwen3.6-plus} call per root-to-state path. Record fields and user instructions are embedded with \texttt{text-embedding-v4}, a Qwen3-Embedding model~\citep{zhang2025qwen3embedding}, and the target selector is \texttt{qwen3.5-flash}~\citep{qwen2026qwen35}. The navigation actors are UI-TARS-1.5-7B and Qwen-VL-Max (\texttt{qwen-vl-max-latest}); the closed-loop variant (Appendix~\ref{app:dynamic-variants}) also describes the live screen with Qwen-VL-Max. GPT-4o~\citep{hurst2024gpt4o} is the evaluation judge. No proprietary data leaves the local environment except through these API calls.

\subsection{Dynamic-memory variants}
\label{app:dynamic-variants}
The closed-loop variant localizes the current screen against the graph at every step and refreshes progress annotations; the plan-edit variant rewrites the memory block when the actor signals confusion. Both are controlled ablations of the canonical open-loop \method\ configuration evaluated in Section~\ref{sec:experiments}.

\end{document}